\def\cvprPaperID{0000}
\def\confName{CVPR}
\def\confYear{2023}

\def\paperTitle{Emotional Speech-Driven Animation with Content-Emotion Disentanglement}

\def\authorBlock{
Radek Daněček$^1$\\
{\tt\small rdanecek@tue.mpg.de}
\and
Kiran Chhatre$^{2}$ \\
{\tt\small chhatre@kth.se}
\and
Shashank Tripathi$^1$\\
{\tt\small stripathi@tue.mpg.de}
\and
Yandong Wen$^1$\\
{\tt\small ywen@tue.mpg.de}
\and
Michael J. Black$^1$\\
{\tt\small black@tue.mpg.de}
\and
Timo Bolkart$^1$\\
{\tt\small tbolkart@tue.mpg.de}
\and
$^1$Max Planck Institute for Intelligent Systems, 
T{\"u}bingen, Germany
\and
$^2$KTH, 
Stockholm, Sweden
}

\newif\ifreview 
\newif\ifarxiv \newcommand{\arxiv}{\arxivtrue}
\newif\ifcamera 
\newif\ifrebuttal  

\newcommand{\model}{EMOTE\xspace}
\newcommand{\modellong}{\model~(Expressive Model Optimized for Talking with Emotion)}
\newcommand{\prior}{FLINT\xspace}
\newcommand{\priorlong}{\prior~(\textbf{FL}AME \textbf{IN} \textbf{T}ime)}

\newcommand{\todelete}[1]{}
\newcommand{\camrdy}[1]{\textcolor{olive}{#1}}
\newcommand{\moved}[1]{\textcolor{brown}{#1}}
\newcommand{\correction}[1]{\textcolor{blue}{#1}}

\arxiv

\newif\ifcamready
\camreadytrue 

\ifcamready
\renewcommand{\camrdy}[1]{#1}
\renewcommand{\moved}[1]{#1}
\renewcommand{\correction}[1]{#1}
\renewcommand{\todelete}[1]{}
\fi

\pdfoutput=1
\documentclass[10pt,twocolumn,letterpaper]{article}
\ifreview \usepackage[review]{cvpr} \fi
\ifarxiv \usepackage[pagenumbers]{cvpr} \fi
\ifrebuttal \usepackage[rebuttal]{cvpr} \fi
\ifcamera \usepackage{cvpr} \fi

\usepackage{graphicx}
\usepackage{amsmath}
\usepackage{amssymb}
\usepackage{booktabs}
\usepackage[normalem]{ulem}

\usepackage{times}
\usepackage{microtype}
\usepackage{epsfig}
\usepackage[table,xcdraw]{xcolor}
\usepackage{caption}
\usepackage{float}
\usepackage{placeins}
\usepackage{color, colortbl}
\usepackage{stfloats}
\usepackage{enumitem}
\usepackage{tabularx}
\usepackage{xstring}
\usepackage{multirow}
\usepackage{xspace}
\usepackage{url}
\usepackage{subcaption}
\usepackage{xcolor}
\usepackage{afterpage}
\usepackage[hang,flushmargin]{footmisc}

\usepackage{graphicx}
\usepackage{amsmath}
\usepackage{amssymb}
\usepackage{comment}
\usepackage{booktabs} 
\usepackage{numprint}
\usepackage{balance}

\ifcamera \usepackage[accsupp]{axessibility} \fi

% ...

\let\shortcite\cite

\ifarxiv  \fi

\newcommand{\R}[1]{{%
    \textbf{%
        \ifstrequal{#1}{1}{\textcolor{red}{R#1}}{%
        \ifstrequal{#1}{2}{\textcolor{blue}{R#1}}{%
        \ifstrequal{#1}{3}{\textcolor{magenta}{R#1}}{%
        \ifstrequal{#1}{4}{\textcolor{teal}{R#1}}{%
                           \textcolor{cyan}{R#1}%
        }}}}%
    }%
}}

\usepackage{xr-hyper}

\makeatletter
\newcommand*{\addFileDependency}[1]{
  \typeout{(#1)}
  \@addtofilelist{#1}
  \IfFileExists{#1}{}{\typeout{No file #1.}}
}

\makeatother

\usepackage[pagebackref,breaklinks,colorlinks]{hyperref}
\usepackage[capitalize]{cleveref}
\crefname{section}{Sec.}{Secs.}
\crefname{table}{Table}{Tables}
\crefname{figure}{Fig.}{Figs.}

\begin{document}
\title{\paperTitle}
\author{\authorBlock}

\twocolumn[{
\renewcommand\twocolumn[1][]{#1}
\maketitle
\begin{center}
    \centering
    \vspace{-0.25in}
    \includegraphics[width=1.99\columnwidth]{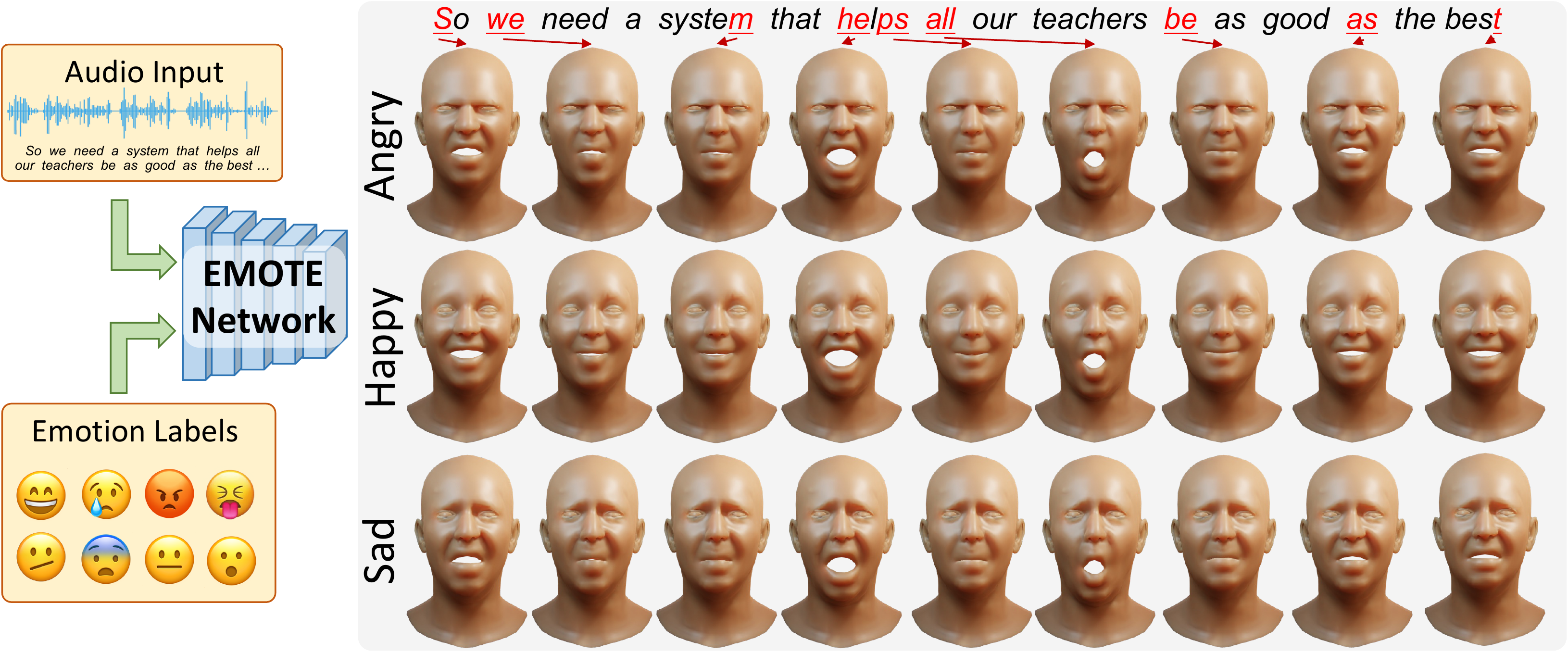}
    \vspace{-0.2cm}
    \captionof{figure}{
        Given audio input and an emotion label, \model generates an animated 3D head that has state-of-the-art lip synchronization while expressing the emotion. 
        The method is trained from 2D video sequences using a novel video emotion loss and a mechanism to disentangle emotion from speech.
    }
    \label{fig:teaser}
\end{center}

}] 

\newcommand{\vect}[1]{\mathbf{#1}}

\newcommand{\norm}[1]{\left\lVert#1\right\rVert}

\newcommand{\shapecoeff}{\boldsymbol{\beta}}
\newcommand{\shapedim}{{\left| \shapecoeff \right|}}
\newcommand{\shapespace}{\mathcal{S}}
\newcommand{\shapespaceexpl}{\mathbb{R}^{\shapedim}}
\newcommand{\posecoeff}{\boldsymbol{\theta}}
\newcommand{\posedim}{{\left| \posecoeff \right|}}
\newcommand{\posespace}{\mathcal{P}}
\newcommand{\posespaceexpl}{\mathbb{R}^{\posedim}}
\newcommand{\jawpose}{\boldsymbol{\theta}_{jaw}}

\newcommand{\jawposerec}{\boldsymbol{\widehat{\theta}}_{jaw}}
\newcommand{\expcoeffrec}{\boldsymbol{\widehat{\psi}}}

\newcommand{\expcoeff}{\boldsymbol{\psi}}
\newcommand{\expdim}{{\left| \expcoeff \right|}}
\newcommand{\expspace}{\mathcal{E}}
\newcommand{\expspaceexpl}{\mathbb{R}^{\expdim}}
\newcommand{\numverts}{n_v}
\newcommand{\numfaces}{n_f}
\newcommand{\template}{\textbf{T}}

\newcommand{\numjoints}{k}
\newcommand{\joints}{\textbf{J}}
\newcommand{\jointregressor}{\mathcal{J}}
\newcommand{\blendweights}{\mathcal{W}}
\newcommand{\blendweightsdim}{\left| \mathcal{W} \right|}

\newcommand{\landmark}{\textbf{k}}


\newcommand{\lighting}{\textbf{l}}
\newcommand{\cam}{\textbf{c}}

\newcommand{\albedo}{A}
\newcommand{\albedocoeffs}{\boldsymbol{\alpha}}
\newcommand{\albedodim}{\left| \albedocoeffs \right|}
\newcommand{\albedospaceexpl}{\mathbb{R}^{\albedodim}}
\newcommand{\normalcoeffs}{\boldsymbol{\nu}}
\newcommand{\normaldim}{\left| \normalcoeffs \right|}

\newcommand{\uvsize}{d}
\newcommand{\image}{I}

\newcommand{\flamev}{M}
\newcommand{\normals}{N}

\newcommand{\facerec}{R}
\newcommand{\spectre}{\facerec_s}
\newcommand{\emoca}{\facerec_e}

\newcommand{\valence}{v}
\newcommand{\arousal}{a}
\newcommand{\expression}{\mathbf{e}}

\newcommand{\asrnet}{A}
\newcommand{\audio}{\mathbf{w}}
\newcommand{\speechfeat}{\mathbf{s}}
\newcommand{\decoder}{D}

\newcommand{\style}{S}
\newcommand{\cond}{\boldsymbol{c}}





\newcommand{\coarsedecoder}{R_c}
\newcommand{\renderer}{R}

\newcommand{\videmonet}{E_{\textup{emo}}^{\textup{vid}}}
\newcommand{\emonet}{E_{\textup{emo}}^{\textup{im}}}
\newcommand{\emovec}{\boldsymbol{\epsilon}}
\newcommand{\emovecdim}{\left|\emovec\right|}
\newcommand{\emometric}{d_e}

\newcommand{\emovidnet}{V}
\newcommand{\emovidvec}{\boldsymbol{\phi}}
\newcommand{\videmovec}{\emovidvec}
\newcommand{\videmovecdim}{\left|\videmovec\right|}
\newcommand{\latentmean}{\boldsymbol{\mu}}
\newcommand{\latentsigma}{\boldsymbol{\sigma}}
\newcommand{\latent}{\boldsymbol{z}}
\newcommand{\latentratio}{q}
\newcommand{\tmp}{^{1:T}}
\newcommand{\tmpl}{^{1:T/\latentratio}}

\newcommand{\motionenc}{\textup{ENC}}
\newcommand{\motiondec}{\textup{DEC}}


\newcommand{\lipnet}{E_{\textup{lip}}}
\newcommand{\lipvec}{\boldsymbol{\eta}}
\newcommand{\lipmetric}{d_l}


\newcommand{\motiondim}{d}

\newcommand{\secref}[1]{Sec.~\ref{#1}}
\newcommand{\figref}[1]{Fig.~\ref{#1}}
\newcommand{\tabref}[1]{Tab.~\ref{#1}}
\newcommand{\supmat}{Sup.~Mat.\xspace}
\begin{abstract}
To be widely adopted, 3D facial avatars \todelete{need to}\camrdy{must} be animated easily,
realistically, and directly from speech signals.
While the best recent methods generate 3D animations that are synchronized with the input audio, they largely ignore the impact of emotions on facial expressions.
Realistic facial animation requires lip-sync together with the natural expression of emotion.
To that end, we propose \modellong, which generates 3D talking-head avatars that maintain lip-sync from speech while enabling explicit control over the expression of emotion.
\todelete{Due to the absence of high-quality aligned emotional 3D face datasets with speech, \model is trained from an emotional video dataset (i.e., MEAD).}
\todelete{To train \model, we match speech content between generated sequences and target videos differently from emotion content.}
\camrdy{To achieve this, we supervise \model with decoupled losses for speech (i.e., lip-sync) and emotion.}
\camrdy{These losses are based on two key observations: (1) deformations of the face due to speech are spatially localized around the mouth and have high temporal frequency, whereas (2) facial expressions may deform the whole face and occur over longer  
intervals.
Thus we train \model with a \camrdy{per-frame} lip-reading loss to preserve the speech-dependent content, while supervising emotion at the sequence level.}
Furthermore, we employ a content-emotion exchange mechanism in order to supervise different emotions on the same audio, 
while maintaining the lip motion synchronized with the speech.
To employ deep perceptual losses without getting undesirable artifacts, we devise a motion prior in the form of a temporal VAE.
\moved{
Due to the absence of high-quality aligned emotional 3D face datasets with speech, \model is trained \camrdy{with 3D pseudo-ground-truth extracted} from an emotional video dataset (i.e., MEAD).
}
Extensive qualitative\todelete{, quantitative,} and perceptual evaluations demonstrate that \model 
produces \correction{speech-driven facial animations with better lip-sync than state-of-the-art methods trained on the same data}, 
while offering additional, high-quality emotional control. 

\end{abstract}

\section{Introduction}
\label{sec:intro}

Animating 3D head avatars solely from speech has numerous applications,
including character animation in films and games, 
virtual telepresence for AR and VR, and the embodiment of digital personal assistants. 
For the best user experience, the speech-driven animation methods must be speaker-independent (i.e., generalize to the audio and facial geometry of unseen subjects) and the lip articulation must be synchronized with the speech content. 
Lip-sync of the generated animation with the audio has drawn significant  attention \cite{cudeiro2019capture, richard2021meshtalk, fan2022faceformer, bala2022imitator, xing2023codetalker}.
But to be lifelike, facial animations must also \camrdy{express natural emotions through facial expressions.}
The modeling of facial expressions in 3D has also been well studied \cite{3DMM_survey}.
What is missing, however, is the modeling and animation of {\em emotion during speech}.

The core issue is that the 3D face shapes that convey emotion are often inconsistent with the lip motions needed to realistically match the audio. 
That is, there is a conflict between expressing the emotion and synchronizing the lips with the audio.
To address this limitation, we present \modellong, a speech-driven 3D facial animation method with semantic animation control over the expressed emotion\todelete{ of the generated animation}.
\model addresses the core problem, making 3D animations that convey the appropriate emotion without hurting lip-sync possible.

Emotional speech in the context of 3D facial animation has been previously ignored due to the absence of suitable datasets. 
While existing speech-driven animation methods are trained on 3D scan datasets with paired audio such as VOCASET \cite{cudeiro2019capture}, BIWI \cite{eth_biwi_00760}, or Multiface \cite{wuu2022multiface}, no large-scale 3D scan dataset with emotional speech sequences exists. 
For this reason, we train \model on MEAD \cite{kaisiyuan2020mead}, an emotional {\em video} dataset, which does not provide 3D supervision.
To \camrdy{compensate for the lack of 3D data}, we generate pseudo ground-truth (GT) 3D data using a combination of \todelete{the best} state-of-the-art (SOTA) monocular reconstruction methods \cite{MICA:ECCV2022, Feng2021_DECA, EMOCA:CVPR:2021, filntisis2022visual} fine-tuned on MEAD.

Directly training a speech-driven animation model following an architecture of a SOTA method such as FaceFormer \cite{fan2022faceformer} on such \camrdy{pseudo-GT}\todelete{ video data}, however, results in mediocre motions, where speech-dependent and emotion-dependent lip articulations are poorly disentangled. 
Furthermore, FaceFormer does not \camrdy{enable any control over} the output emotions.

We observe that facial animation is the result of two factors that differ temporally and spatially: speech and emotional state.
Specifically, speech-induced articulation (or lip-sync) \camrdy{has a high temporal frequency; the lip shape must match the audio at every point in time. }
In contrast, emotions are a longer-lasting phenomenon that change at a 
\camrdy{lower temporal frequency compared with speech-driven articulation.}
Additionally, speech production is localized around the mouth region, whereas facial expressions may occur over the entire face region.
We hypothesize that these temporal and spatial differences make it possible to disentangle these two phenomena.

Specifically, in order to enforce the consistency of the lip-sync, we apply a per-frame lip-reading consistency loss similar to \cite{filntisis2022visual} while enforcing the desired emotion at the sequence level through a novel transformer-based dynamic emotion consistency loss.
Finally, to separate the effect of emotion from the effect of the spoken words, we propose a novel emotion-content disentanglement mechanism that we use to train our model.

Naively training a SOTA network such as FaceFormer with the aforementioned components leads to temporally unstable and unnatural results. 
To ensure that the generated motion is natural and temporally consistent, we first train a facial motion prior, specifically a temporal transformer-based VAE that operates over sequences of 3DMM (FLAME \cite{FLAME:SiggraphAsia2017}) parameters. 
We then train a regressor to map the 
speech audio onto the latent space of the 
prior.


With this, \model generates high-quality 3D facial animations with accurate lip-sync while enabling the editing of the expressed emotion. 
We demonstrate the ability to edit emotions qualitatively and quantitatively in perceptual studies. 


Our contributions are summarized as: 
(1) The first method for semantic emotion editing of speech-driven 3D facial animation.
(2) A novel supervision mechanism with perceptual lip-reading and dynamic emotion losses and a novel content-emotion disentanglement mechanism. 
(3) 
A statistical prior for facial motion that is designed to support manipulation of facial motion with perceptual losses while keeping the animation natural.
\todelete{(4) An efficient feed-forward (non-autoregressive) transformer network }
\camrdy{(4) A bidirectional non-autoregressive architecture that is more efficient than autoregressive transformer-based SOTA methods.}
\todelete{(5) The first method that regresses into a parametric face model (FLAME).} 
The pseudo ground-truth 3D \camrdy{(FLAME parameters)} for the MEAD dataset, the trained \model, and code to train and generate speech-driven facial animations with emotion control \camrdy{are available} for research purposes at 
\ifcamready
\url{https://emote.is.tue.mpg.de/}
\fi

\section{Related Work}
\label{sec:related}

%
The field of speech-driven 3D facial animation \todelete{also} has a long history \cite{cao2005expressive, edwards2016jali, edwards2020jalicyberpunk, xu2013practical, taylor2012dynamic, cohen2001animated}.
We focus on the most relevant recent work, which leverages deep learning \cite{pham2017end, pham2017speech, karras2017audio, taylor2017generalized, zhou2018visemenet, cudeiro2019capture, richard2021meshtalk, fan2022faceformer, xing2023codetalker}. 

\paragraph{Semantic Control:}
Few methods provide the user with any kind of semantic control of the generated 3D avatar.
VOCA \cite{cudeiro2019capture} and FaceFormer \cite{fan2022faceformer} allow the speaking style to be controlled by interpolating the style vectors of training individuals;  this does not enable simple editing of emotion.  \
While 
MeshTalk \cite{richard2021meshtalk} can generate a variety of results for the same audio input, there is no mechanism that allows for any control of emotion.
Karras et al.~
\shortcite{karras2017audio} 
learn a type of emotional latent space by jointly learning a feature vector for each training sample in an unsupervised way. 
Changing this feature vector then allows test-time editing. 
The learned space, however, does not inherently contain a semantic meaning and this must be manually assigned after training.
Since there is no disentanglement mechanism, the model lacks the guarantee that mixing different emotion vectors with different audio input will produce the desired result (i.e.~correct lip-sync and desired emotion).
Concurrent, and most relevant to \model is EmoTalk~\cite{Peng2023_EmoTalk}, a method to animate emotional 3D faces from speech input. 
Unlike \model, EmoTalk requires artist-curated training data, and it only provides  control over the intensity of the expression of emotion, but not over the emotion type.
In contrast, \model is, to the best of our knowledge, the first to factor the effect of emotion and speech on the resulting 3D animation via a novel emotion-content disentanglement mechanism, allowing semantically meaningful emotion editing at test time.

\camrdy{
Works such as \cite{edwards2016jali, taylor2017generalized, zhou2018visemenet} automatically animate artist-controllable FACS rigs but also lack explicit speech-driven emotion control and some require additional inputs, i.e., a transcript \cite{edwards2016jali, edwards2020jalicyberpunk}.
JALI for Cyberpunk \cite{edwards2020jalicyberpunk} shows characters with emotional faces, however, the amount of artist work, manually designed rules, and hand-crafted features needed to build the system is unclear.
}

\paragraph{Supervision:}
The recent methods are fully supervised \cite{karras2017audio, cudeiro2019capture, richard2021meshtalk, fan2022faceformer, bala2022imitator, xing2023codetalker}, requiring a training dataset of 3D scans paired with the synchronized speech. 
Notably, these methods use a mean squared error loss between the predicted and ground truth mesh vertices (or vertex offsets from a template mesh) at each frame.
Richard et al.~\shortcite{richard2021meshtalk} introduce a cross-modal loss to enforce reconstruction of audio-correlated and uncorrelated information separately in order to learn a categorical motion prior and an explicit eye-closure loss to enforce eye blinks.
Thambiraja et al.~\shortcite{bala2022imitator} introduce a mouth-closure loss that is active only when bilabials are spoken, which helps achieve proper mouth closure.
\model goes further to use perceptual lip- and emotion-consistency losses with a novel disentanglement framework.


\paragraph{Motion prior:}
Most methods do not learn or apply any type of motion prior and let the space of valid motions be learned by the architecture itself from the training data \cite{pham2017end, pham2017speech, karras2017audio, cudeiro2019capture, fan2022faceformer}.
MeshTalk \cite{richard2021meshtalk}, on the other hand, adopts a two-stage approach. In the first stage, a motion prior with a categorical latent space is trained. 
This pretrained prior is then used in the second stage to autoregressively generate the results. 
CodeTalker \cite{xing2023codetalker} adopts \todelete{a similar} \camrdy{an} approach \camrdy{similar} to FaceFormer \cite{fan2022faceformer} and augments it with a separately trained VQ-VAE motion prior.
\camrdy{
Chandran et al.~\shortcite{chandran2022facial} introduce a transformer-based autoencoder for facial motion animation with disentangled identity and shape. The authors demonstrate its effectiveness for tasks like motion compression, retargetting, unconditional generation and others. 
However, its suitability for regression tasks like speech-driven animation is not investigated.
}

\paragraph{2D talking head generation:}
There is a long line of work focused on generating 2D videos of talking heads given speech \cite{AverbuchElor, tripathy2020icface, Zakharov2019, kim2019neural, chen2019hierarchical, choi2020starganv2, Ding2017ExprGANFE, suwajanakorn2017obama, lindt19, prajwal2020wav2lip, dApolito_2021_CVPR, ji2021evp, cheng2022VideoReTalking, Tripathy_2021_facegan}
and today there are even commercial systems for this task.
These approaches, however, typically ignore emotion and focus on lip-sync.
The few methods that address facial-expression animation operate over 2D videos ~\cite{tripathy2020icface,paraperas2022ned,ji2021evp, thies2020nvp}.
While some of these methods use 3D parametric models to guide the output expressions (e.g., \cite{paraperas2022ned}), their focus is not on outputting the 3D shape and, hence, the underlying 3D shapes are of low quality.
%
%
%
%

\paragraph{3D Datasets:}
The ideal dataset for our task would 
contain ground-truth (GT) 3D face scans synchronized with audio. 
Such data is limited due to the expense and complexity involved in capturing it.
BIWI \cite{eth_biwi_00760}, VOCASET~\cite{cudeiro2019capture}, S3DFM~\cite{10.1016/j.sigpro.2019.02.025}, and Multiface~\cite{wuu2022multiface} are publicly available datasets for the audio-driven 3D talking-head task. 
These datasets are limited in size, richness of emotion, speaking styles, and shapes of the subjects.


\paragraph{2D Datasets:}
In contrast to the limited richness of 3D datasets, 2D video is plentiful.
Specifically, there are many available video-speech datasets \cite{nagrani2017voxceleb,Chung2018_VoxCeleb2,roessler2018faceforensics,zhu2022celebv,afouras2018deep,afouras2018lrs3,kaisiyuan2020mead}, 
and video datasets focused speech emotion recognition (SER) \cite{livingstone2018ryerson,cao14_crema,poria2018meld,cmumosi,cmumosei}.
See the \supmat for an overview of existing 2D and 3D datasets.
Of the existing video datasets, MEAD \cite{kaisiyuan2020mead} is most suitable for our task. 
It is of sufficient size, it is captured in the lab, which makes it easier to perform 3D face reconstruction than in-the-wild video, and, most importantly, it exhibits high emotional variety. 

Off-the-shelf 3D face-reconstruction methods can be applied to the video frames, providing pseudo-GT data.
This, however, comes with many drawbacks. 
While the field of image-based 3D face reconstruction has made tremendous progress \cite{Deng2019,Genova2018,Tewari2017,Tewari2018,Tewari2019_FML, Sanyal2019_RingNet,Feng2021_DECA,Shang2020_MGCNET,yang2020facescape, EMOCA:CVPR:2021, filntisis2022visual, MICA:ECCV2022, zollhoefer_survey_2018}, SOTA methods are often not robust to occlusion, they produce inaccurate shape or expression, or are not temporally stable. 
Despite these limitations, the large amount of data available from video outweighs the downsides.
Consequently, we generate pseudo-GT data from video by integrating recent SOTA methods \cite{EMOCA:CVPR:2021, MICA:ECCV2022, filntisis2022visual}.

\section{Background and Notation}

\ifcamready
\begin{figure*}
    \offinterlineskip
    \centerline{    
    \includegraphics[width=1.99\columnwidth]{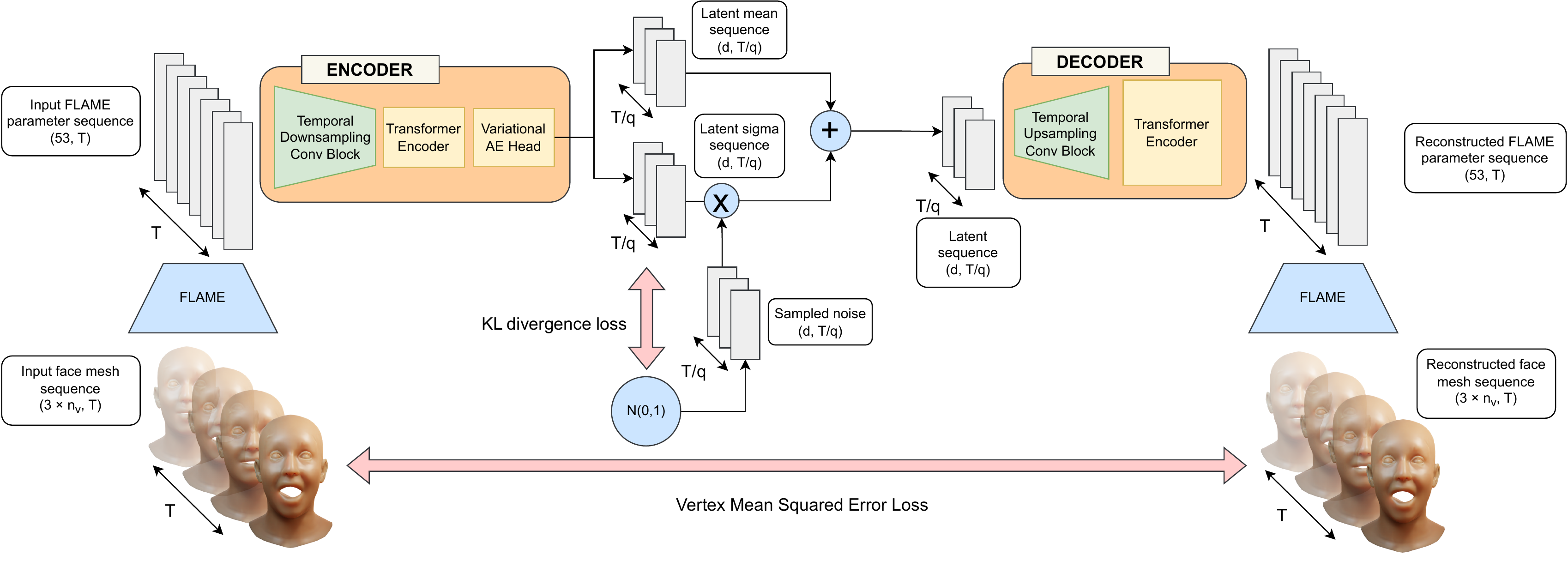}
    }
    \caption{
        \textbf{\prior motion prior architecture.}
        Given a sequence of $T$ FLAME parameters, the encoder maps this sequence to a sequence of compact latents. The decoder then reconstructs this latent sequence back into a sequence of $T$ FLAME parameters. The reparametrization trick is employed to sample the latents from predicted sequences of means and sigmas.
    }
    \label{fig:prior_arch}
\end{figure*}
\fi

\label{sec:prelim}
\paragraph{Face model:}
\model predicts the expression and jaw pose parameters of FLAME~\cite{FLAME:SiggraphAsia2017}, a parametric 3D head model. 
FLAME is defined as a function $\flamev(\shapecoeff, \posecoeff, \expcoeff)\rightarrow (\mathbf{V}, \mathbf{F})$  that takes parameters for identity shape $\shapecoeff \in \mathbb{R}^\shapedim$, facial expression $\expcoeff \in \mathbb{R}^\expdim$, and pose $\posecoeff \in \mathbb{R}^{3\numjoints+3}$, with $\numjoints=4$ joints, and outputs a 3D mesh with vertices $\mathbf{V} \in \mathbb{R}^{\numverts \times 3}$ and triangles $ \mathbf{F} \in \mathbb{R}^{\numfaces \times 3}$.

\paragraph{Emotion feature extraction:}
We use EMOCA's~\cite{EMOCA:CVPR:2021} publicly available emotion recognition network to predict emotion features from an image.
This model consists of a ResNet-50 \cite{He2016_ResNet} network trained on AffectNet \cite{affectNet} on the task of in-the-wild emotion recognition to regress
valence and arousal and classify the eight basic expressions (neutral, happiness, sadness, surprise, fear, disgust, anger, and contempt). 
After training the network, the prediction head is discarded, and the output of the final ResNet layer is used as an emotion feature vector, $\emovec \in \mathbb{R}^{\emovecdim}$
.
In the following, we denote the network with $\emonet(\image) \rightarrow  \emovec$.

\paragraph{Video emotion feature extraction:}
\camrdy{Emotions are phenomena that generally last at least several seconds or more. } 
Single
-frame emotion features are insufficient to describe 
\camrdy{them,}
\camrdy{because an expression in a single frame carries the effect of both emotion and speech. This can lead to}
misinterpreting speech-induced articulation for emotional cues \camrdy{(see Fig.~3 in \supmat).} 
\camrdy{Hence, }emotion features \todelete{for videos must} \camrdy{should} aggregate information across time. 
To address this, we train a lightweight transformer-based emotion classifier that takes a sequence of emotion features $ \emovec \tmp \in \mathbb{R}^{T \times \emovecdim}$ and outputs a video emotion classification vector $\expression \in \mathbb{R}^{8}$ and the video emotion feature $\videmovec \in \mathbb{R}^{\videmovecdim}$, which is the sequence-aggregated feature produced by the last transformer layer before the classification head, 
\camrdy{with $\videmovecdim=256$.}
We refer to the video motion feature extraction as $\videmonet ( \emovec \tmp ) \rightarrow \left( \expression, \videmovec \right)$.
\camrdy{More details about the video emotion extraction can be found in \supmat}

\paragraph{Speech feature extraction:}
To encode the audio signal, we employ a pretrained ASR network, Wav2Vec 2.0 \cite{baevski2020wav2vec}. 
It takes as input the raw waveform sampled at 16kHz. 
This waveform is first passed through temporal convolutional layers producing a feature sampled at 50Hz. 
Similar to Fan et al.~\shortcite{fan2022faceformer}, we use linear interpolation to downsample the feature down to 25Hz to match the frame-rate of our input videos. 
The resampled feature is then fed into the transformer-based part of Wav2Vec 2.0, producing the output speech feature. 
Formally, it is defined as $\asrnet(\audio) \rightarrow \speechfeat \tmp$, 
where $\asrnet$ is the Wav2Vec 2.0 network, $\audio$ is the raw waveform, and 
$ \speechfeat \tmp \in \mathbb{R}^{T \times d_s}$ 
is the final speech feature resampled to 25Hz. $T$ denotes the number of frames and each frame is of dimension $ d_s=768 $.
\section{Method}
\label{sec:method}
\paragraph{Motivation:}

\ifcamready
\begin{figure*}[t]
    \offinterlineskip
    \centerline{    
    \includegraphics[width=1.99\columnwidth]{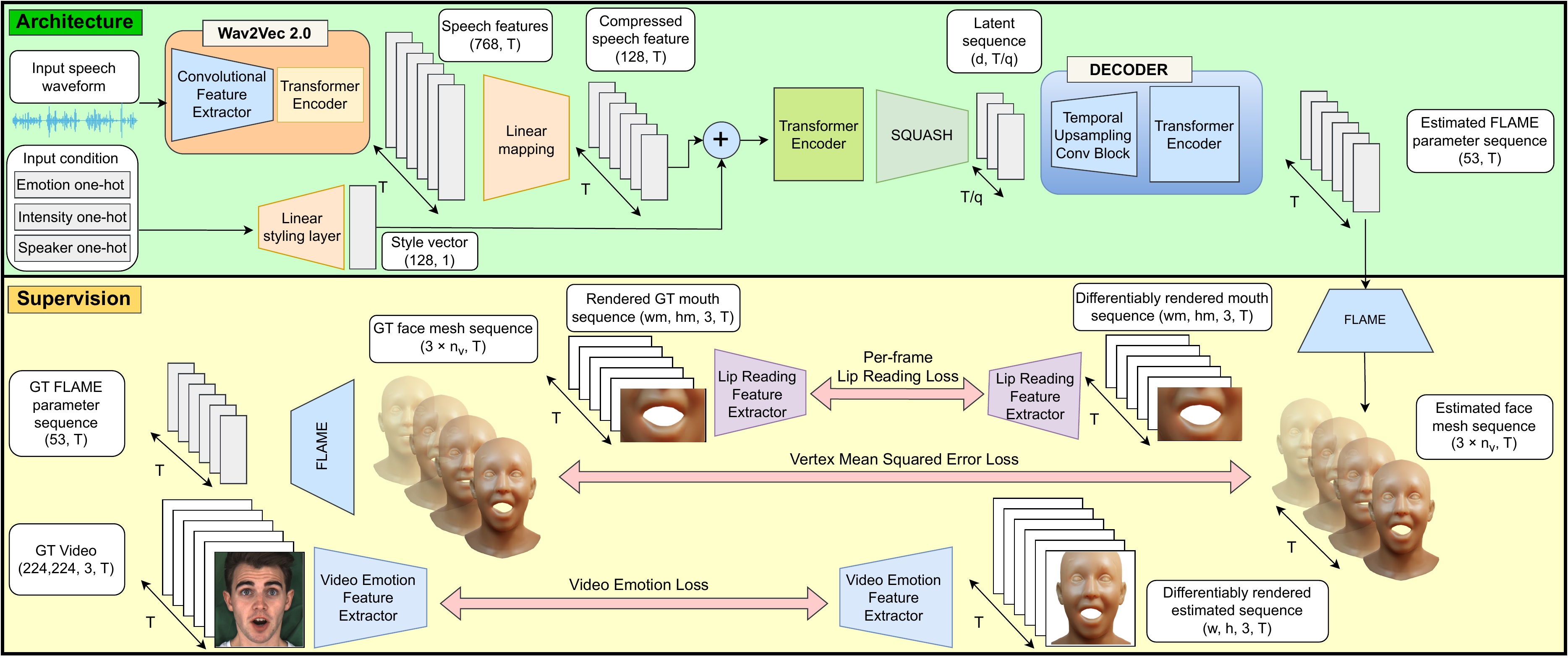}
    }
    \caption{
         \model architecture. The speech input consists of the raw audio waveform (top left) and conditioning, which includes one-hot vectors of the training speaker ID and, importantly, the emotion class and intensity. The audio is encoded by Wav2Vec 2.0 and the input conditions are mapped with a linear styling layer. These two are then concatenated and passed through additional convolutional 
         layers to map down to the latent space sequence. The \camrdy{pretrained frozen} \prior decoder transforms this sequence back into a FLAME parameter sequence. Finally, the meshes are generated by the FLAME forward pass.
    }
    \label{fig:a2f_arch}
\end{figure*}
\else
\fi

\model follows a two-step pipeline, which first trains a temporal variational autoencoder, and then uses its latent space as a motion prior. 
Specifically, we train a regressor that maps the speech audio to the latent space of the motion prior conditioned on a given target emotion, its intensity (mild, medium, or high), and a subject-specific speaking style.

\subsection{Facial Motion Prior: \prior}

\label{subsec:prior}
Facial motion is complex \camrdy{and modeling it is challenging. 
To simplify the problem we represent it in a 
learned low-dimensional representation.}
As a foundation, we represent the face in each of the $T$ frames of a sequence using FLAME, giving 
$\expdim +|\jawpose|=53$ 
dimensions (i.e., 50 expression parameters and 3 jaw pose parameters) per frame.
Facial motions, however, are not independent between frames and, hence, a sequence can be represented in a lower-dimensional space.
To that end, we train a temporal variational autoencoder called \priorlong\/ to represent facial motion sequences.
The formulation exploits a transformer encoder to extend the VAE framework to our temporal modeling problem (cf.~\cite{ng2022learning2listen}).
We exploit this as a prior in training \model and find that it reduces high-frequency jitter and unnatural jaw rotations.

\paragraph{Architecture:}
The encoder compresses the sequence \camrdy{of $T$ frames 
$(\expcoeff \tmp, \jawpose \tmp )$}
\camrdy{into} $T/\latentratio$ latent frames $ \latent \tmpl $, 
where $\latentratio$ is the number of consecutive 
\camrdy{original} frames that a single latent frame encapsulates \camrdy{(similar to \cite{ng2022learning2listen})}.
\camrdy{The intervals of consecutive latents do not overlap.}
We empirically set $\latentratio=8$. 
Specifically,
 \begin{equation}
     \motionenc(\expcoeff \tmp, \jawpose \tmp ) \rightarrow (\latentmean \tmpl, \latentsigma \tmpl).
 \end{equation}
Using the VAE reparametrization 
gives us the final latent sequence: 
$
    \latent^t = \latentsigma^t * \latent_s^t + \latentmean^t, 
$
where $\latent^t$ is one latent frame and $\latent_{s}^t$ is sampled from 
$ \mathcal{N}(\mathbf{0},\mathbf{I}) $. 
This is done separately for each latent frame $t \in \{1, ..., T / \latentratio \} $, 
before they are stacked to compose the final sequence of latents, $ \latent \tmpl $.
The sequence  $ \latent \tmpl $ is then decoded back to the original space:
\begin{equation}
    \motiondec(\latent \tmpl ) \rightarrow ( \widehat{\expcoeff} \tmp, \widehat{\posecoeff}_{jaw} \tmp ), 
    \label{eq:motion_decoder}
\end{equation}
The architecture of our autoencoder is outlined in 
\ifcamready
Fig.~\ref{fig:prior_arch}, 
\fi 
and the hyperparameters are in \supmat

\paragraph{Losses:}
We train \prior with the following loss functions: 
\begin{equation} 
L_{\textup{total}} = \lambda_{\textup{rec}}L_{\textup{rec}} + \lambda_{\textup{KL}}L_{\textup{KL}} .
\end{equation}

\noindent
\paragraph{Reconstruction loss:}
For each frame $t$,
we compute the mean squared error between the pseudo-GT and predicted meshes: 
\begin{equation}
    L_{\textup{rec}} = \textup{MSE}(\mathbf{\widehat{V}} \tmp, \mathbf{V} \tmp), 
\end{equation}
where the vertex coordinates $\mathbf{V}^t, \mathbf{\widehat{V}}^t $ are produced by feeding the GT and reconstructed parameters through FLAME: 
%
$
     \mathbf{V}^t  = \flamev(\shapecoeff, \expcoeff^t, \jawpose^t).
$

\noindent
\paragraph{KL divergence:}
For each latent frame in the sequence, we compute the standard VAE KL divergence term \cite{kingma2013vae}. 
\begin{equation}
    L_{\textup{KL}} = 0.5\left[-\sum_i(\log\sigma_i^2 + 1) + \sum_i\sigma_i^2 + \sum_i\mu^2_i\right] .
\end{equation}
Please note that the individual latent means and sigmas are not treated here as a sequence but separately.

\subsection{Emotional Speech-Driven Animation: \model}
\paragraph{Architecture:} 

\model is an encoder-decoder architecture summarized in Fig.~\ref{fig:a2f_arch}. 
The encoder uses Wav2Vec 2.0 \cite{baevski2020wav2vec} to extract the audio feature sequence: $\asrnet(\audio) = \speechfeat \tmp$.
Each extracted audio feature $\speechfeat^t$ is concatenated with a style vector: 
\todelete{. 
The style vector is a linear projection of the conditioning emotion, intensity, and speaker style input $\cond$: } 
$
     \speechfeat_s \tmp= \left[ \style(\cond) \tmp | \speechfeat \tmp \right],
$
\camrdy{with $\style(\cond)$ denoting the styling function, which is a linear projection of the input condition $\cond$.}
At training time, $\cond$ is the 
\camrdy{ground-truth}
emotion \camrdy{type}, \camrdy {emotion} intensity, and speaker ID: 
\begin{equation}
    \camrdy{
    \cond = \left[\cond_{emo} | \cond_{int} | \cond_{id} \right],
    }
    \label{eq:cond}
\end{equation}
\camrdy{where $\cond_{emo}, \cond_{int}, \cond_{id}$ are one-hot vectors of emotion, intensity and identity indices.}
At test time, 
$\cond$ can be set manually, which provides animator control over the emotion of the output sequence. 
After the style is incorporated, the speech feature is mapped to the latent space of the  motion prior. 
Specifically, it is temporally downsampled by concatenating $q$ consecutive frames together and then projecting it with a linear layer down to a single latent frame:
$
    \textup{SQUASH}(\speechfeat_s \tmp) = \latent \tmpl.
$
Finally, the obtained latent sequence $ \latent \tmpl$ is fed to the pretrained, frozen, motion decoder to produce the output FLAME parameters using Eq.~\ref{eq:motion_decoder}, obtaining the estimates of 
$\widehat{\expcoeff} \tmp, \widehat{\posecoeff}_{jaw} \tmp$.

During training, both the GT and predicted geometry are rendered with a differentiable renderer \cite{Ravi2020_PyTorch3D}, and the images are passed to a lip-reading network $\lipnet$ and video emotion network $\videmonet$.
We denote the forward pass, including the differentiable rendering, and the extraction of emotion and lip-reading features as: 
\begin{equation}
          \textup{\model}( \speechfeat\tmp, \cond ) \rightarrow ( \widehat{\mathbf{V}} \tmp,  \widehat{\lipvec} \tmp , \widehat{\videmovec} ),
\end{equation}
where $\widehat{\mathbf{V}} \tmp$ is the 
\camrdy{generated} vertex sequence, $\widehat{\lipvec} \tmp $ is the sequence of lip-reading features, and $\widehat{\videmovec}$ is the video emotion feature.

\camrdy{Note that unlike recent transformer-based SOTA methods, \model is not autoregressive. Hence, the decoder is only called once, making the decoding computational complexity $O(1)$. This is more efficient than the $O(T)$ autoregressive decoding loop of FaceFormer and CodeTalker. It also allows us to consider future context by employing bidirectional decoding similar to BERT \cite{devlin2018bert}.}

\paragraph{\camrdy{Training:}}
During training, we 
supervise the model using the following loss functions: 
\begin{equation} 
L_{\textup{total}} = L_{\textup{rec}} +  L_{\textup{emo}} + L_{\textup{lip}}  +  L_{\textup{emo}}^{\textup{dis}} + L_{\textup{lip}}^{\textup{dis}}.
\end{equation}

\paragraph{Reconstruction loss:}
For each frame $t$ in the sequence we compute the mean squared error between the pseudo-GT and predicted meshes: 
\begin{equation}
    L_{\textup{rec}} = \textup{MSE}(\mathbf{\widehat{V}} \tmp, \mathbf{V} \tmp).
\end{equation}

\paragraph{Video emotion loss:}
We extract the video emotion feature from the original video 
$\videmonet(\emonet(I\tmp)) = \videmovec$ 
and from the differentiably-rendered predicted sequence and call this
$\widehat{\videmovec}$. 
Their emotional content should be the same, so we penalize their distance:
\begin{equation}
    L_{\textup{emo}} = \emometric(\widehat{\videmovec}, \videmovec ),
\end{equation}
where $\emometric$ is the negative cosine similarity.

\paragraph{Lip-reading loss:}
For each frame $t$ in the sequence, we also compute a perceptual lip-reading loss.
We crop out the mouth region and feed it to the lip-reading network. 
We extract per-frame lip-reading features using $\lipnet$ and calculate the distance between the 
pseudo-GT
lip-reading features and the predicted lip-reading features: 
\begin{equation}
    L_{\textup{lip}} = \lipmetric(\widehat{\lipvec} \tmp, \lipvec \tmp),
\end{equation}
where $\lipmetric$ is the negative cosine similarity. 


\paragraph{\moved{Disentangling  emotion and content:}}
\begin{figure}[t]
    \offinterlineskip
    \centering
    \includegraphics[width=0.99\columnwidth]{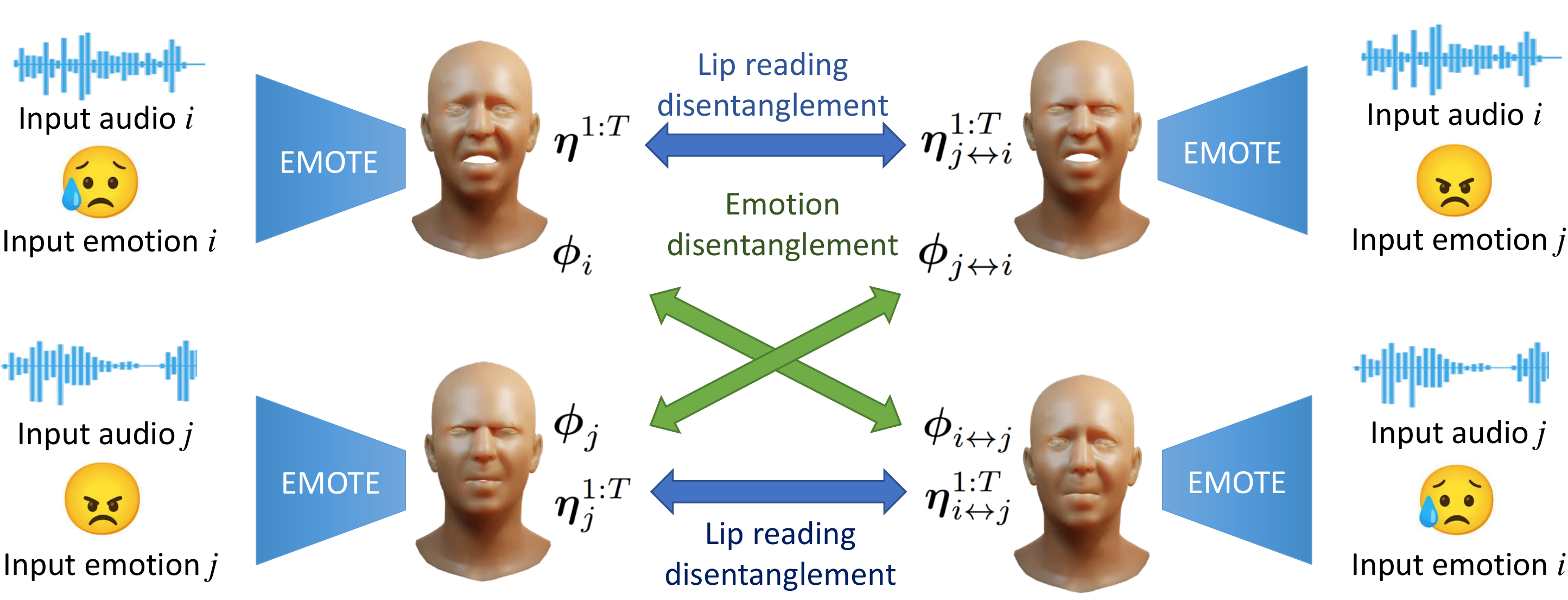}
    \caption{
        Disentanglement mechanism. During training, we duplicate batches and exchange the emotion condition in the duplicated batch (right), the augmented batch is also passed through the model, and we compute the disentanglement losses such that the result of the augmented batch keeps the original articulation but takes on the desired emotion.
    }
    \label{fig:disentangle}
\end{figure}
\moved{
Our goal is to disentangle content and emotion such that we can  
control one while retaining the other. 
The conditioning described 
\camrdy{in Eq.~\ref{eq:cond}} is not sufficient to achieve this. 
Hence, we devise a novel emotion-content disentanglement mechanism, outlined in Fig.~\ref{fig:disentangle}.
During training, we take two sequences with different emotions and switch their emotion conditions.
The lip-reading loss for each decoded result should match the original despite the change in emotion, but the emotion in the decoded result should change to match the new condition.
}

\moved{
More formally, let $ \textup{\model}( \speechfeat_{i} \tmp, \cond_{i} ) = ( \widehat{\mathbf{V}}_i \tmp,  \widehat{\lipvec}_i \tmp , \widehat{\videmovec}_i ) $ be a forward pass of \model for sample $i$ and similarly for a distinct index $j$,  $\textup{\model}( \speechfeat_{j} \tmp, \cond_{j}) $.
We also generate sequences with swapped emotion conditions, i.e.,
$ \textup{\model}( \speechfeat_{i} \tmp, \cond_{j\leftrightarrow i}) = 
\camrdy{( \widehat{\mathbf{V}}_{j\leftrightarrow i} \tmp,  \widehat{\lipvec}_{j\leftrightarrow i} \tmp , \widehat{\videmovec}_{j\leftrightarrow i} )} $ 
and 
$ \textup{\model}( \speechfeat_{j} \tmp, \cond_{i\leftrightarrow j}) = 
\camrdy{( \widehat{\mathbf{V}}_{i\leftrightarrow j} \tmp,  \widehat{\lipvec}_{i\leftrightarrow j} \tmp , \widehat{\videmovec}_{i\leftrightarrow j} )} 
$, \camrdy{with $i\!\!\leftrightarrow\!\!j$ denoting generations using audio $j$ with the emotion and intensity condition of audio $i$.
}
}
%

\paragraph{Disentanglement losses:}
We apply both emotion and lip-reading perceptual losses to the augmented samples: 
\begin{align}
    L_{\textup{emo}}^{\textup{dis}} = \emometric(\widehat{\videmovec}_{i\leftrightarrow j}, \videmovec_i) & &
    L_{\textup{lip}}^{\textup{dis}} = \lipmetric(\widehat{\lipvec}_{i\leftrightarrow j} \tmp, \lipvec_j \tmp).
\end{align}
\todelete{
where $i, j$ are indices connecting GT emotion feature and lip-reading features $ \videmovec_i,  \lipvec_j \tmp$ and the corresponding disentangled predicted features $\tilde{\videmovec}_i, \tilde{\lipvec}_j \tmp$. 
Note that $\tilde{\emovec}_i$ is the result of generating using audio $i$ and emotion $j$, i.e., the swapped condition. 
}
\camrdy{
Since we treat emotion as a sequence phenomenon, rather than a per-frame phenomenon, we can bypass the requirement for temporal alignment between emotion features of $\videmovec_{i\leftrightarrow j} $ and $ \videmovec_i$. 
}

\section{Implementation Details}
\label{sec:details}

\paragraph{Data:}
\todelete{The motion prior and audio2face }
\camrdy{\prior and \model}
are trained on the MEAD dataset~\cite{kaisiyuan2020mead}. 
MEAD is an emotional video dataset of 48 subjects, each uttering around 30 short English sentences. 
\todelete{Each subject talks in a neutral emotion and seven basic emotions} 
\camrdy{Each subject utters all sentences several times, once for neutral and three times for seven basic emotions}
(i.e., anger, disgust, fear, happiness, contempt, sadness, and surprise), where each of the basic emotions is articulated with three intensity levels. 
The subjects are actresses and actors fluent in English. 
We 
\camrdy{use 39 subjects and}
split the dataset such that \todelete{70\%}\camrdy{32} \todelete{of the }subjects are included in the training set \todelete{, 15\%}\camrdy{and 7} in the validation set\todelete{, and 15\%  in the test set}; i.e., training\todelete{,} \camrdy{and} validation\todelete{, and testing} sets use different subjects.

For evaluation, we use audio sequences from LRS3~\cite{afouras2018lrs3}, a large scale dataset of English TED and TEDx talks by a large variety of speakers.
We use the LRS3 test set in our evaluations, which is disjoint from the speakers in our training set. 

\paragraph{Data processing:}
Since MEAD does not come with 3D meshes, we recover the 3D faces synchronized with the audio directly from the videos. 
However, we found that existing monocular 3D face reconstruction methods (e.g., \cite{Feng2021_DECA, EMOCA:CVPR:2021, filntisis2022visual}) are insufficient to process the data.
Specifically, EMOCA~\cite{EMOCA:CVPR:2021} best recovers emotional 3D faces, but it is often over-expressive and does not well match the lip articulation.
SPECTRE~\cite{filntisis2022visual} improves the lip articulation but lacks expressiveness of the emotion.
Both methods use DECA \cite{Feng2021_DECA} to recover shape, but this is less accurate than MICA~\cite{MICA:ECCV2022}. 
To get the best of these methods, we augment EMOCA with SPECTRE's lip-reading loss, replace its predicted identity face shape with MICA's prediction, and use Mediapipe~\cite{mediapipe} keypoints instead of FAN~\cite{Bulat2017} keypoints as supervision. 
We then finetune the image encoders of this combined model on MEAD by minimizing EMOCA's losses with an additional keypoint loss, and the added lip articulation loss. 
Once fine-tuned, running inference of the refined model on MEAD gives the reference 3D face shapes. 
\camrdy{More details can be found in \supmat}

\paragraph{Motion prior:}
\prior is trained on the MEAD dataset for 500 epochs with batch size 4 \todelete{, a}\camrdy{and} sequence lengths of 32 frames.
Adam \cite{Kingma2015} is used as optimizer, with a learning rate of 1e-4. 
The size of the latent space is set empirically to 128 and $q = 8$, and the KL divergence term is weighted with a factor of 1e-3. 



\paragraph{Speech-driven animation model:}
We train \model in two stages. 
In the first stage, we only supervise with the vertex loss  without the disentanglement mechanism. 
This is computationally efficient since it does not require differentiable rendering. 
We train the model for 20 epochs with batch size 4 and sequence lengths of 64 frames with the Adam optimizer and a learning rate of 1e-4. 
The first stage is only supervised using the MSE of vertex differences with weight: $\lambda_{\textup{rec}}=1$. 
In the second stage, we freeze the wav2vec weights, add the differentiable rendering, enable the perceptual losses and the disentanglement mechanism, and fine-tune for two more epochs.
The perceptual loss weights are: $ \lambda_{\textup{emo}} = \lambda_{\textup{emo}}^{\textup{dis}} = 2.5e-6$ 
and $\lambda_{\textup{lip}} = \lambda_{\textup{lip}}^{\textup{dis}} = 2.5e-5$. 

\section{Experiments}
\label{sec:experiments}

Evaluation
must consider two components: the sync \todelete{of}\camrdy{between} the lip articulation and the input speech and the quality of the emotional content.
As both tasks are difficult to evaluate automatically, we \camrdy{conduct} two perceptual experiments \moved{and provide qualitative evaluations} 
to demonstrate the quality and effectiveness of \model .

\subsection{Perceptual Studies}

We conduct two perceptual studies on Amazon Mechanical Turk. 
First, we compare \model's lip-sync quality with that of publicly available SOTA methods. 
Second, individual model components are ablated in order to evaluate the influence of each component on the perceived quality of the results.

\paragraph{Lip articulation evaluation:}
This study compares \model with 
CodeTalker \cite{xing2023codetalker}, FaceFormer \cite{fan2022faceformer}, MeshTalk \cite{richard2021meshtalk}, and VOCA \cite{cudeiro2019capture}.
We randomly selected 15 input audio sequences from the LRS3 test set and used these to synthesize the facial motion. 
For results generated with \model, the input emotion condition was set to neutral.
In the study, we showed the participants two audible output videos of two different methods side by side (the left-right order was randomized for every hit). 
The participants played both videos separately. 
After playing both videos at least once, the participant was \todelete{ asked}\camrdy{allowed} to select the result with better lip-sync on a 5-point Likert scale (strong/weak preference for one or the other model, or equally good).
Each of the two-way comparison studies was completed by 15 participants. 
Three catch trials with obvious answers (videos with animation generated by a different audio than the one playing) were added, and participants that preferred the catch trials were 
excluded (see figures for details).

\todelete{Figure~\ref{fig:perceptual_sota} shows histograms of} 
\camrdy{The Likert plot in Fig.~\ref{fig:perceptual_sota} shows}
preferences averaged across participants. 
Note that all SOTA methods were trained on high quality audio-4D scan datasets (\cite{cudeiro2019capture, wuu2022multiface}), while \model is trained on pseudo-GT (i.e., MEAD). 
\todelete{
Regardless, \model's lip-sync is perceived as better than most methods and on-par with CodeTalker.
}
\correction{
\model's lip-sync is preferred over scan-trained MeshTalk despite being trained on pseudo-GT. 
Note that the lip-sync of methods trained on VOCASET (VOCA, FaceFormer, CodeTalker) is preferred due to the superior training data quality of VOCASET's 3D scans.
}
\todelete{Further, }
\correction{Importantly, in a fair comparison,}
\model outperforms FaceFormer retrained on MEAD, suggesting the value of our architecture and method.

\paragraph{Ablation experiments:}
This study evaluates the importance of the individual building blocks of \model.
We compare: 
(1) \model,
(2) \model w/o the disentanglement terms,
(3) \model w/o disentanglement and emotion loss,
(4) \model w/o disentanglement and lip-reading loss,
(5) \model w/o \prior,
(6) \model w/o disentanglement and perceptual losses,
(7) \model w/ static emotion loss computed per-frame instead of the dynamic emotion loss,
(8) FaceFormer-EMO, which is the FaceFormer architecture augmented with a one-hot input for the emotion condition and intensity.

For this study, we randomly select 14 input audios from the LRS3 test set and use these to synthesize the facial motion. 
We ensure that each of the 7 basic emotions \todelete{(happy, sad, angry, fear, surprise, disgust, and contempt)}\camrdy{(anger, disgust, fear, happiness, contempt, sadness and surprise)} is equally represented. 
Similar to the study above, we present the participants with two videos. 
First the videos are muted, and the participant is asked which of the two videos better communicates a particular specified emotion.
Then, the same videos are presented but this time with audio and the participant is asked which of the two has better lip-sync (same question as in the study above). 
Participants answer both questions on a 5-point Likert scale. 
This process repeats for all 14 video pairs.
Each of the two-way comparisons was completed by 15 participants. 
As above, three catch trials for both emotion and lip-reading were used to automatically filter out uncooperative participants. 
Figure \ref{fig:perceptual_ablation} reports the results of this study for both emotion and lip-sync. 
The study demonstrates that all \todelete{of the }design choices are critical. 
Perhaps the only surprising result is the similar performance of an \model variant that uses a static per-frame emotion loss instead of the dynamic one. 
For more details, see the visual ablation study.

\subsection{Qualitative Results}
\label{subsec:qualitative}

\begin{figure}[t]
    \offinterlineskip
    \centering
    \includegraphics[width=1.0\columnwidth]{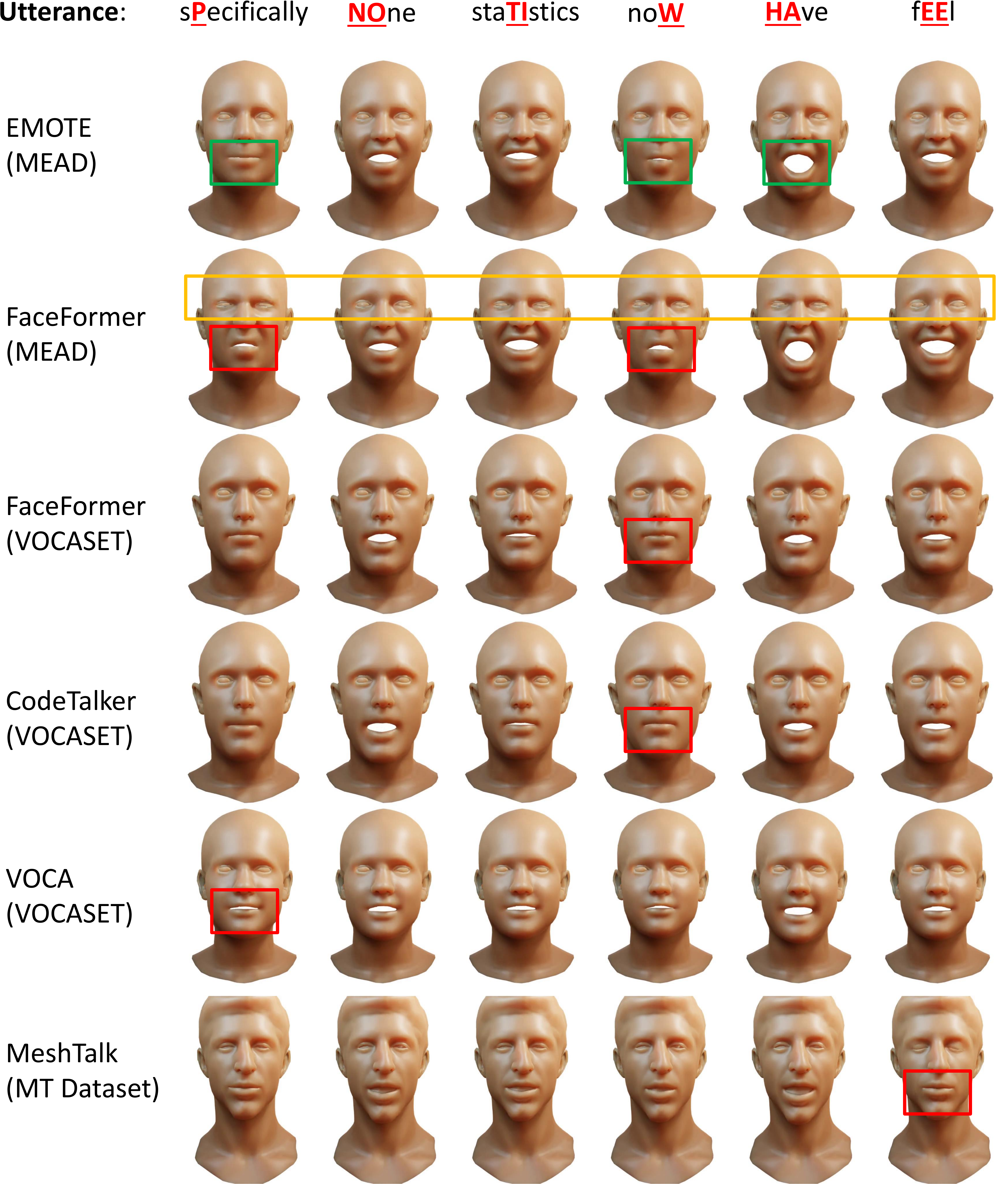}
    \caption{
        \textbf{Visual comparison to SOTA}. 
        The rows show specific frames from a ``neutral" sequence generated by EMOTE and baseline models in order: FaceFormer (trained on MEAD pseudo-GT), FaceFormer, CodeTalker, VOCA, and MeshTalk. 
        \camrdy{The training dataset of each method is indicated in the brackets.}
        The input characters of the utterances that were used to generate the animations are highlighted in red (top text description). 
        Our model generates accurate animations of the lips 
        \camrdy{(highlighted in green)}. 
        FaceFormer trained on the pseudo-GT MEAD data, without emotion condition, \camrdy{struggles} to produce a consistent ``neutral" emotion \camrdy{given the neutral input speech} (highlighted orange). Baseline models\camrdy{, while giving mostly good results,}  \camrdy{can sometimes also } exhibit inferior lip-sync (highlighted in red).
    }
    \label{fig:compare_sota}
\end{figure}


\paragraph{Comparison with SOTA:}
In Fig. \ref{fig:compare_sota}, we qualitatively compare \model with the SOTA methods. 
While all methods produce good lip articulation in accordance with the spoken words, none of these methods is able to produce emotional animations. 
\todelete{While} FaceFormer trained on our MEAD training data can produce emotional faces, \camrdy{however}, speech-content and emotions are not well disentangled, and method lacks emotional control.
This highlights the importance of our emotion-content disentanglement mechanism.

\paragraph{Ablation experiment:}

\begin{figure*}[t]
    \offinterlineskip
    \centering
    \includegraphics[width=1.80\columnwidth]{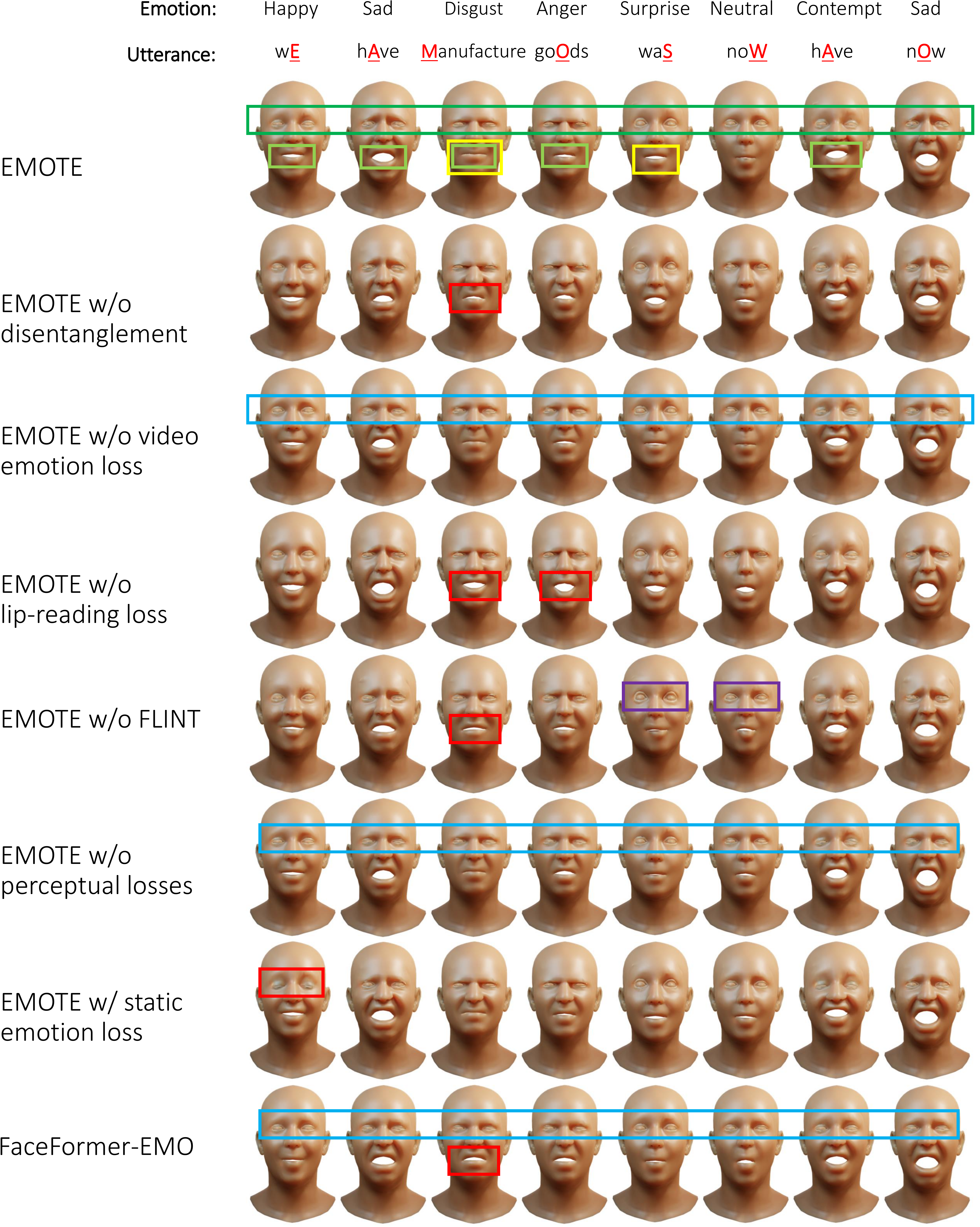}
    \caption{
        \textbf{Visual ablation study}. 
        The rows show generations of the specified model for several spoken words highlighted above. 
        Here we include different visemes and emotions taken out of context (not in sequence).
        Our model exhibits high emotional fidelity (highlighted green) and accurate lip motions \camrdy{(yellow)}. 
        \camrdy{The effects of emotion are also visible in the lower part of the face (light green).}
        Models without emotional supervision suffer from poor emotional fidelity (highlighted blue). 
        Models without explicit lip-reading supervision often suffer from inferior lip-sync (such as incomplete mouth closure on bilabials).
        Finally, the model without \prior yields uncanny artifacts (purple). 
        \camrdy{Please watch the supplementary video to see the results in motion.}
    }
    \label{fig:visual_ablation}
\end{figure*}
\begin{figure*}[h]
    \offinterlineskip
    \centering
   \ifarxiv 
   \includegraphics[scale=0.5]{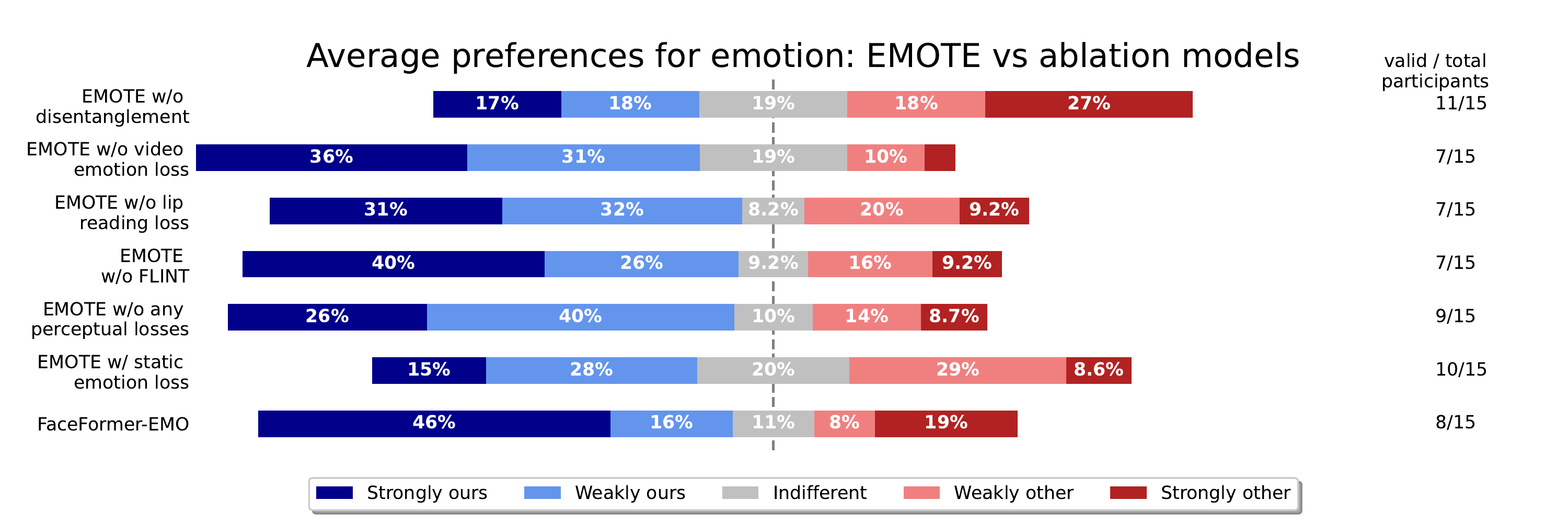}
   \includegraphics[scale=0.5]{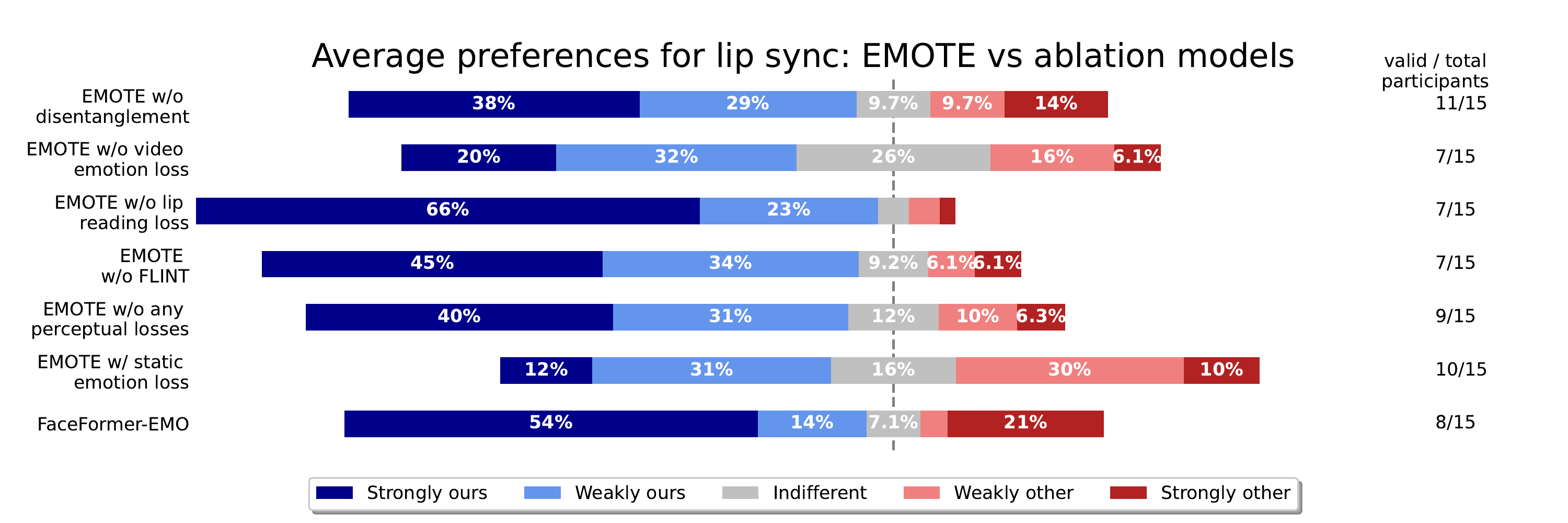}
   \else
   \includegraphics[trim={0.55cm 0 5.2cm 0.5cm},clip, scale=0.355]{figs/perceptual/likert_emo_vs_ablations_.pdf}
   \includegraphics[trim={3.7cm 0 1.55cm 0.5cm},clip, scale=0.355]{figs/perceptual/likert_sync_vs_ablations_.pdf}
   \fi
    \caption{
       \ifarxiv
       Ablation perceptual study results for emotion quality (top) and lip-sync quality (bottom). 
       \else 
       Ablation perceptual study results for emotion quality (left) and lip-sync quality (right). 
       \fi
       \correction{While participants prefer \model w/o disentanglement on the emotion task its inferior articulation hurts the lip-sync preferences (see Fig.~\ref{fig:visual_ablation} and Sup. Video). 
       \model \ /w static emotion loss performs comparably on both metrics but occasionally results in artifacts (see Fig.~\ref{fig:visual_ablation} and Sup. Video).
       \model is preferred on both tasks to all other ablated models.
       }
    }
    \label{fig:perceptual_ablation}
\end{figure*}
\begin{figure*}[t]
    \offinterlineskip
    \centering
    \includegraphics[trim={0.25cm 1.25cm 0.2cm 0.5cm},clip, scale=0.5]{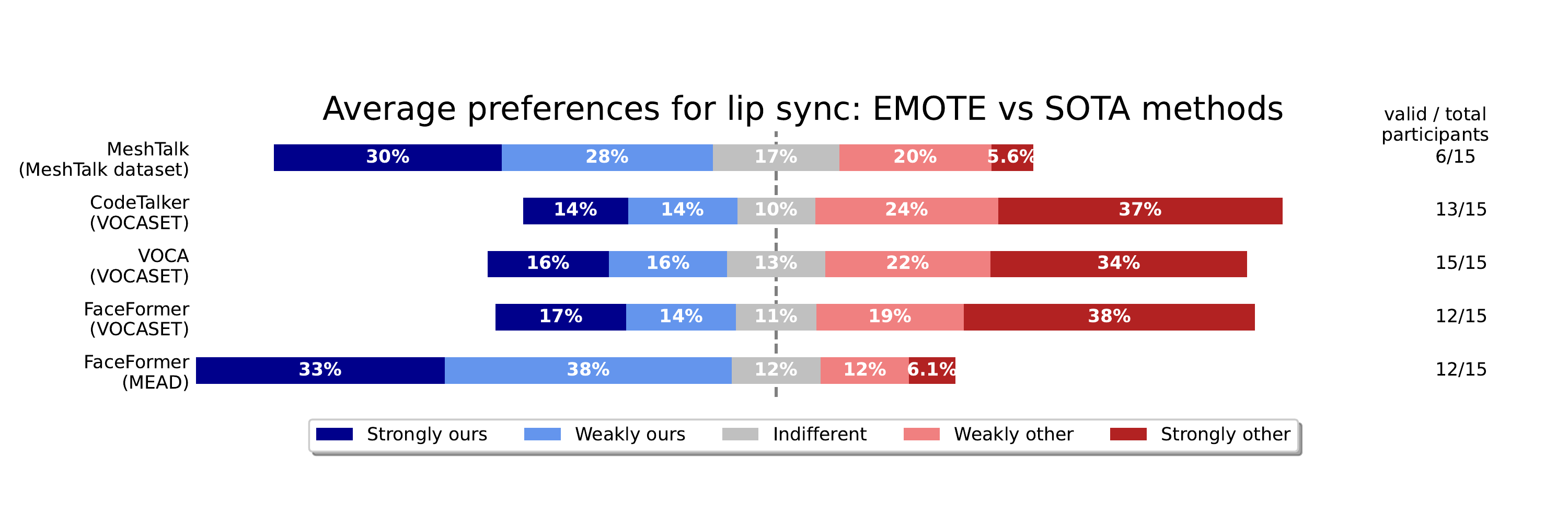} 
    \caption{
        \todelete{The results of the perceptual study comparing lip-sync of our method with the SOTA. The participants prefer \model 
        over MeshTalk }
        \todelete{VOCA}
        \todelete{and FaceFormer} 
        \todelete{trained on MEAD. 
        There is a slight preference over the original FaceFormer.
        Finally, the participants judge \model to be comparable to CodeTalker
        .
        }
        \correction{
        The results of the perceptual study comparing lip-sync of our method with the SOTA. 
        The training dataset of each model is indicated in the brackets.
        The participants prefer \model over MeshTalk. 
        VOCA, FaceFormer and CodeTalker, which are trained on VOCASET, are preferred over \model thanks to the superiority of the scanned training data. 
        However, when we train FaceFormer on MEAD, its lip-sync preference is considerably lower, highlighting the benefits of our approach over SOTA methods on inferior data.
        }
    }
    \label{fig:perceptual_sota}
\end{figure*}

Figure \ref{fig:visual_ablation}  visualizes the effect of the individual design components. 
\model (top row) produces accurate mouth shapes for various words and emotions. 
\model w/o disentanglement terms (second row) starts to lose accurate lip-sync (for instance mouth closures on bilabials), especially during higher intensity emotions. 
\model w/o the video emotion loss (third row) starts to lose some of the emotional cues as the only supervision signal that remains for emotion is through the geometry loss. 
\model w/o the lip-reading loss (fourth row) suffers from inaccurate lip-sync. 
\model w/o \prior is temporally unstable and produces undesirable artifacts. 
\model w/o \todelete{all} any perceptual loss terms 
suffers from lack of emotional expressivity. 
\model with a static emotion loss instead of the dynamic one has results that are comparable to \model but sometimes suffers from undesirable artifacts such as eye closure. 
Finally, FaceFormer-EMO, a variant of the FaceFormer architecture augmented with emotion conditioning of Eq.~\ref{eq:cond}, lacks both expressiveness and accurate lip-sync.







\subsection{Emotion editing}
\begin{figure*}[h]
    \offinterlineskip
    \centering
    \includegraphics[width=2.0\columnwidth]{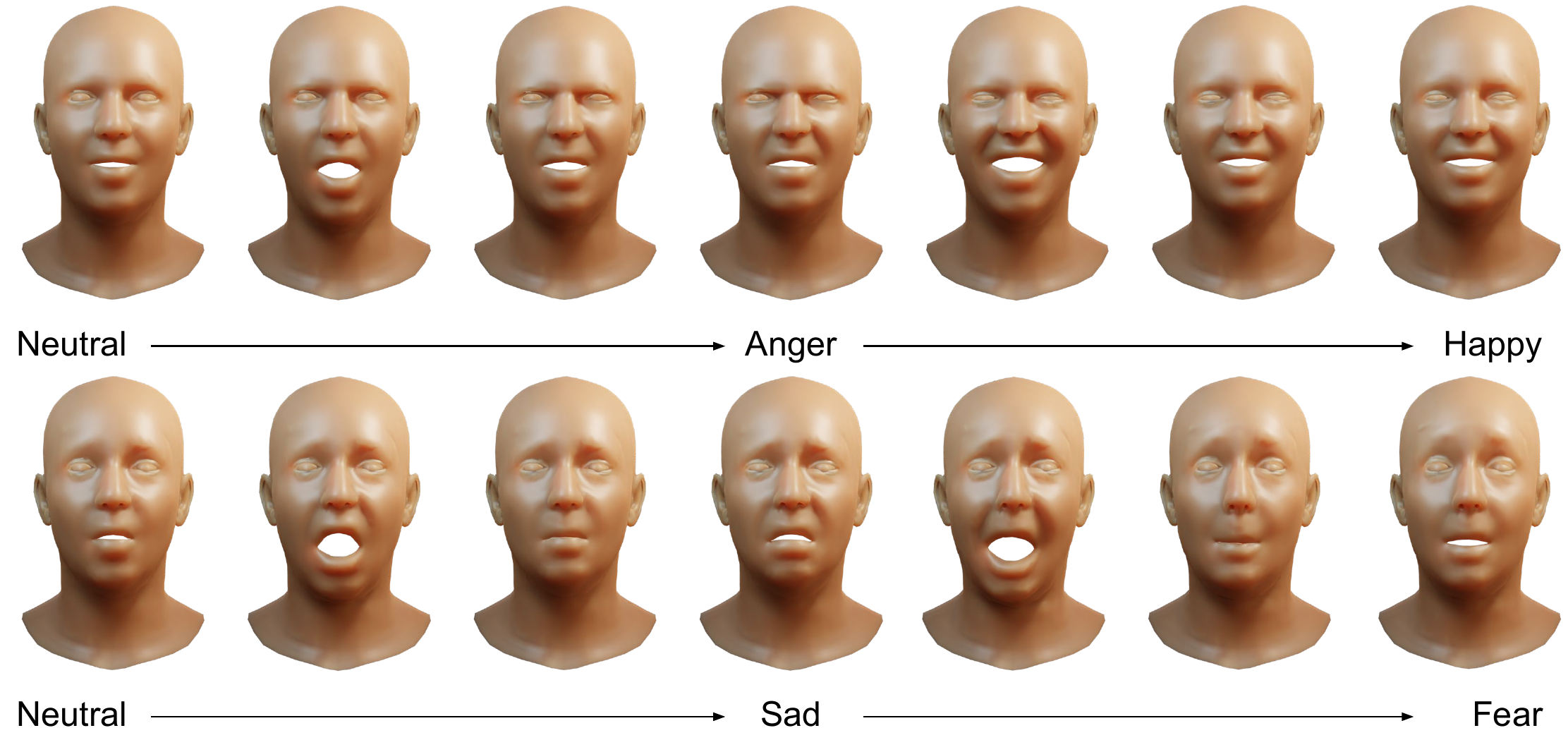}
    \caption{
        \textbf{Emotion editing.} The figure demonstrates that the emotion can be smoothly edited between emotions \textbf{during} speech. 
        Despite the fact that \model is trained with constant emotion labels, it can naturally generalize to edit emotion transitions in a single sequence, thus giving an animator the ability to direct the emotion and its intensity over the course of the animation as they see fit.
    }
    \label{fig:emotion_transitions}
\end{figure*}

\model  provides animators with emotion control\todelete{ over the resulting animation}. Figure \ref{fig:emotion_transitions} demonstrates this effect by editing emotions over the course of a sequence.


\subsection{Limitations}

\paragraph{\camrdy{High frequency speech:}} 
We have demonstrated that \model produces emotional performances while maintaining lip-sync \todelete{comparable} \camrdy{superior} to the SOTA \camrdy{trained on the same pseudo-GT data}.
However, the model is not perfect, as it can fail with very fast high-frequency speech. 
This is due to our data collection process. 
While our pseudo-ground truth has good shapes that capture emotion, it is only sampled at 25fps. 
Using 3D scan data, with a higher sampling rate, \camrdy{or elaborate data augmentation,} could produce more accurate results.

\paragraph{\camrdy{Eye blinks:}}
\todelete{While \model can generate various emotions, it} 
\camrdy{\model}
does not model eye blinks since those are \todelete{not} \camrdy{only weakly} correlated with \todelete{neither} speech \todelete{nor} \camrdy{and} emotion 
\camrdy{and are hence difficult to capture with a deterministic method like \model.} 
\camrdy{Incorporating findings from \cite{ruhland2015eyegaze} may help alleviate this limitation.}

\paragraph{\camrdy{Paralinguistics:}}
\camrdy{
\model does not model paralinguistic motion such as raising eyebrows on words that require emphasis since the lip-reading loss affects only the mouth. 
Solving this may entail incorporating language semantics (i.e., language models), a richer training set and non-deterministic prediction modelling.
}

\paragraph{\camrdy{Emotion granularity:}}
\camrdy{
\model is capable of generating 8 basic emotions in various speaking styles corresponding to the number of training individuals. 
However, realistic emotion-induced motion can take on many more emotions and many more styles. 
Incorporating this would require training on large-scale datasets of sufficient richness and a more granular emotion model.
}

\paragraph{\camrdy{Mouth cavity:}}
\todelete{Finally, }
\model and existing SOTA methods focus on the face shape and ignore the teeth and tongue, which can be important in speech perception.


\paragraph{\camrdy{Automatic emotion control:}}
\camrdy{While \model is capable of producing emotional faces, the emotion label must be provided by the user. 
This process could be automated by using automatic speech emotion recognition to provide the emotion condition.
}



\section{Conclusions}

We have presented \model, the first framework 
to generate 3D talking head avatars with explicit 
control over the type and intensity of emotional expression. 
Unlike current SOTA methods that require high-quality scan datasets for training, \model is trained from an emotional video dataset. 
\todelete{
Despite training on data without high-quality 3D ground truth, \model's lip-sync is perceived on par or preferable compared to existing SOTA. 
}
\camrdy{
Despite training on data without high-quality 3D ground truth, \model's lip-sync is of high quality, and 
better than that of SOTA methods trained on the same data.
}
This is enabled by (co-)supervising the training with perceptual losses, i.e., a video emotion loss and a lip-reading loss, which give \model an edge compared to the SOTA methods that are supervised solely with pseudo-GT geometry.
Without high-quality 3D data, a geometric loss alone is insufficient. 
\model's loss terms ensure that the results carry emotional content as well as accurate lip articulation that is in accordance with the speech signal.
A novel content-emotion exchange mechanism ensures that the lip articulation is driven by the spoken word and the expression is controlled solely by the specified emotion condition, effectively disentangling the two naturally entangled phenomena.
To 
utilize the power of the perceptual losses without artifacts, we devise a temporal transformer-based VAE coined \prior that operates on FLAME parameter sequences. 
We then use its decoder as our motion prior by mapping the speech features and the emotion condition into its latent space.
Unlike the SOTA methods, \model regresses FLAME expression and jaw pose, enabling more direct control over face shape by varying FLAME's identity shape parameters. 
\model makes use of a computationally efficient feedforward architecture.
We believe that \model opens an important and largely overlooked problem in the speech-driven animation field, i.e., that of emotional animation generation, and makes a considerable advance in that direction.

%

\ifcamready
\small{
\paragraph{Acknowledgements:}
\ifarxiv
We thank Alpar Cseke, Taylor McConnell and Tsvetelina Alexiadis for their help with design and deployment of the perceptual study. We also thank Benjamin Pellkofer, Joan Piles-Contreras, Eugen Fritzler and Jojumon Kavalan for cluster computing and IT support. 
Finally, we express our gratitude to Anastasios Yiannakidis, Nikos Athanasiou, Peter Kulits and Maria-Paola Forte for proof-reading and valuable feedback.
\else
We thank A.~Cseke, T.~McConnell and T.~Alexiadis for their help with design and deployment of the perceptual study, B.~Pellkofer, J.~Piles-Contreras, E.~Fritzler and J.~ Kavalan for cluster computing and IT support, and A.~Yiannakidis, N.~ Athanasiou, P.~Kulits and P.~Forte for proof-reading and valuable feedback.
\fi
This project has received funding from the European Union’s Horizon 2020 research and innovation programme under the Marie Skłodowska-Curie grant agreement No.860768 
(CLIPE project).

\paragraph{Disclosure:}

Michael Black has received research gift funds from Adobe, Intel, Nvidia, Meta/Facebook, and Amazon. 
Michael Black has financial interests in Amazon, Datagen Technologies, and Meshcapade GmbH.
While  Michael Black is a consultant for Meshcapade and Timo Bolkart a full-time employee of Google, their research was performed solely at, and funded solely by, the Max Planck Society.
}
\fi

{\small
\bibliographystyle{ieee_fullname}
\bibliography{egbib}
}

\ifarxiv \clearpage \appendix

\section{Additional Technical Details}

\begin{figure*}[t]
    \offinterlineskip
    \centerline{    \includegraphics[width=2.0\columnwidth]{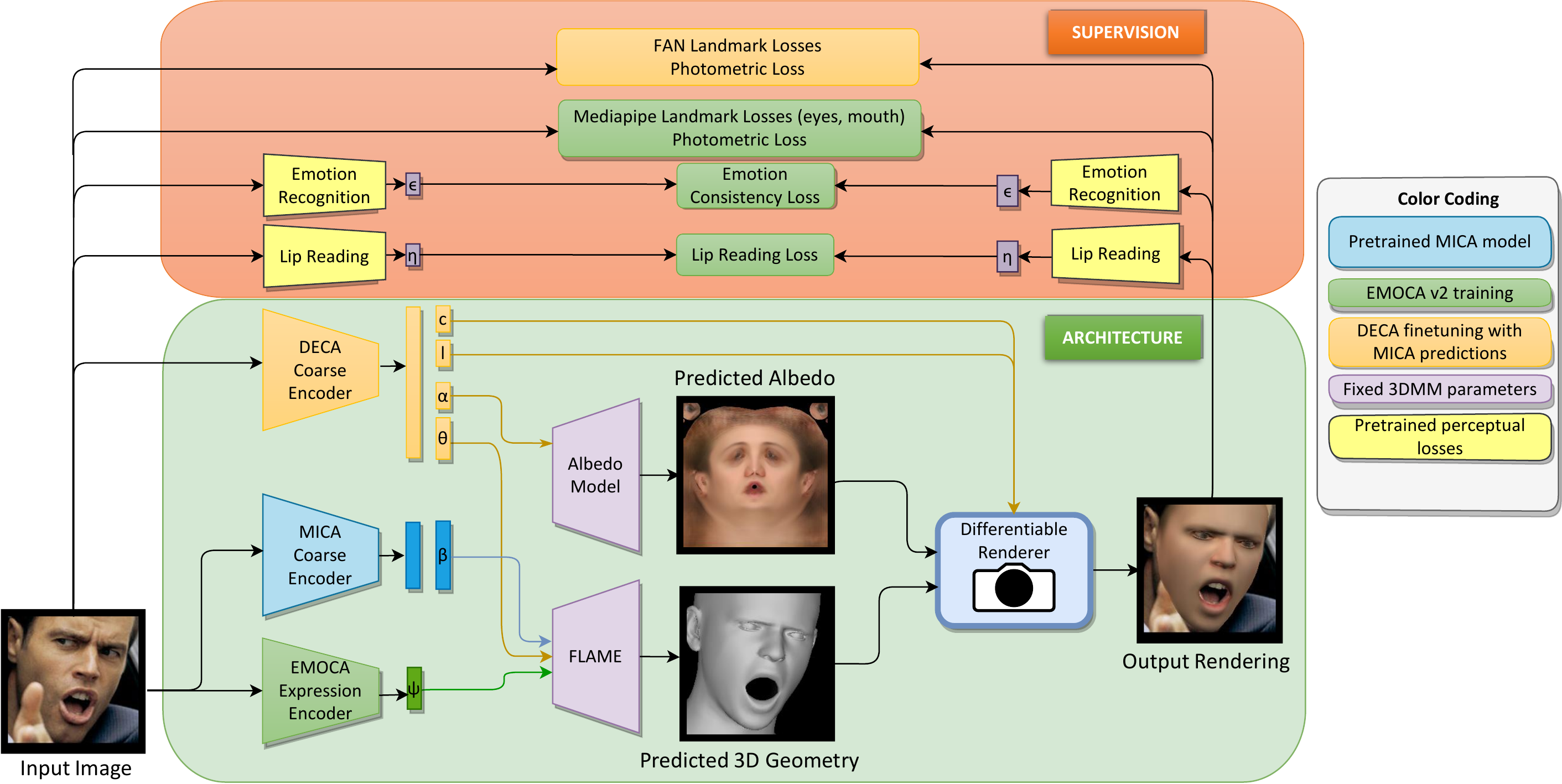}}
    
    \caption{
        \textbf{Face Reconstruction Architecture.}
        The bottom part of the figure shows the architecture components. 
        The top part shows the loss computation. 
        The system is trained in stages, the trainable components of each stage and the corresponding losses are color-coded with the same colors (beige for DECA and green for EMOCA).
    }
    \label{fig:emica_arch}
\end{figure*}

\subsection{Motion Prior: FLINT}

\ifcamready
Here we provide additional details about \prior.
\fi
Both transformers (in encoder and decoder) use a modified version of ALiBi \cite{press2021train}, that allows the transformer to look into the future steps (as opposed to the past steps only, which is what the origina ALiBi does). 
The ALiBi mechanism is preferred over additive positional encodings since it generalizes to arbitrary sequence lengths better \cite{press2021train}.
Both convolutional blocks in encoder/decoder have three convolutional layers that temporally downsample/upsample the sequence. 
The bottleneck dimension is empirically set to $128$.
In training, $\lambda_{\textup{rec}}$ is set to 1000000 and $\lambda_{\textup{KL}}$ to 0.001, which makes the converged KL divergence term less than one order of magnitude lower than the reconstruction terms.
The model code and weights will be made publicly available.

\subsection{Face Reconstruction Network}

In this part we describe our face reconstruction network used to generate pseudo-GT for MEAD. 
In order to obtain the highest quality possible, we employ a combination of four SOTA in-the-wild face reconstruction methods, each of which tackles a particular aspect of the problem. 
Specifically, we augment DECA \cite{Feng2021_DECA} with MICA \cite{MICA:ECCV2022}, the SOTA on face shape prediction.
Next, we utilize the expression prediction of EMOCA \cite{EMOCA:CVPR:2021} to get the SOTA quality of facial expressions and emotions. 
Finally, in training we incorporate the lip-reading loss term from SPECTRE \cite{filntisis2022visual} in order to produce the SOTA-level of lip articulation. 

\paragraph{Architecture:}
The architecture is depicted in Fig.~\ref{fig:emica_arch}. 
The input image is passed through all three encoders (MICA, DECA and EMOCA). 
MICA's encoder is used to output the facial shape vector $ \shapecoeff $. 
DECA's encoder predicts the rest of the parameters: camera $\cam$, spherical harmonics coefficients for lighting $\lighting$, albedo coefficients $ \albedocoeffs $, global head pose and jaw pose $\posecoeff$.  
EMOCA's encoder predicts the facial expression coefficients $\expcoeff$. 
With the regressed predictions, we can now reconstruct the geometry and render an image which can be used for supervision.

\paragraph{Training:}
We finetune the individual components of the aforemention architecture in stages. 

\textbf{(1) Finetuning DECA:}
In the first stage, the DECA \cite{Feng2021_DECA} coarse encoder (beige) is finetuned from the original released DECA model on VGGFace2 \cite{Cao2018_VGGFace2}, the same dataset as the authors. 
The difference from the original implementation is that we take MICA's prediction for the facial shape vector $ \shapecoeff $. 
The encoder of MICA remains frozen.
DECA's encoder predicts the rest of the parameters: $\cam$,  $\lighting$, $ \albedocoeffs $, $\posecoeff$ and also the expression coefficients $\expcoeff$. 

\textbf{(2) Training EMOCA:}
In the second stage, we train the EMOCA expression encoder similarly to  Danecek et al.~\cite{EMOCA:CVPR:2021}, discarding DECA's expression predictions. 
However, there are a few differences. 
Instead of employing FAN \cite{Bulat2017} landmarks, we make use of Mediapipe \cite{mediapipe} landmarks. We only use the landmarks and not the face contour since the face contour does not affect the expression. Compared to FAN landmarks, the eye and mouth landmarks of Mediapipe are more accurate.
In addition to the photometric and emotion consistency loss from the EMOCA authors, we also employ the lip-reading loss from SPECTRE. 
This stage is akin to the EMOCA v2 the authors released but upgraded with MICA for shape prediction. 

\textbf{(3) Finetuning on MEAD:}
In the final stage, we finetune EMOCA with the same losses as in the previous stage on MEAD in order to get the most accurate MEAD pseudo-GT possible.

\subsection{Limitations of Static Emotion Recognition}

\begin{figure*}[t]
    \offinterlineskip
    \centering
    \includegraphics[width=1.\columnwidth]{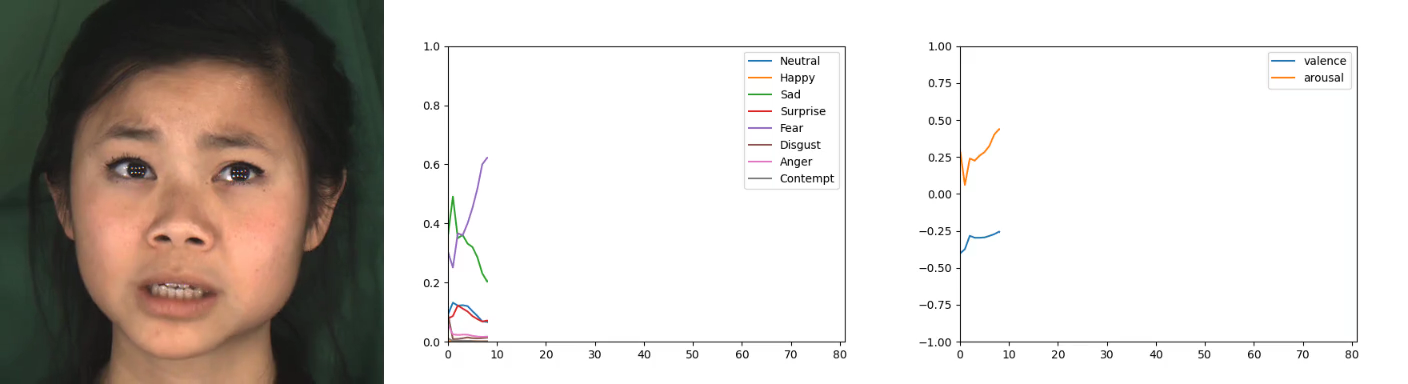}
    \includegraphics[width=1.\columnwidth]{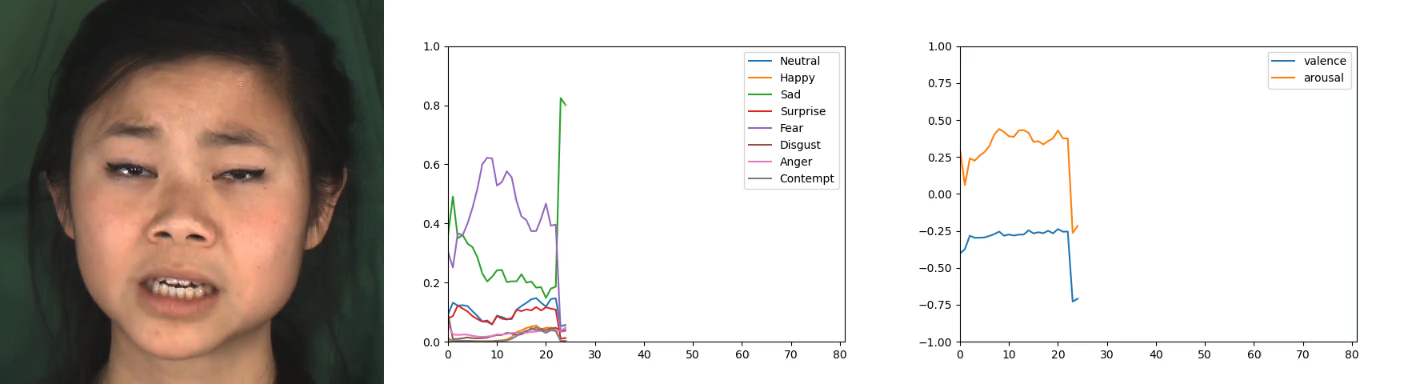}
    \includegraphics[width=1.\columnwidth]{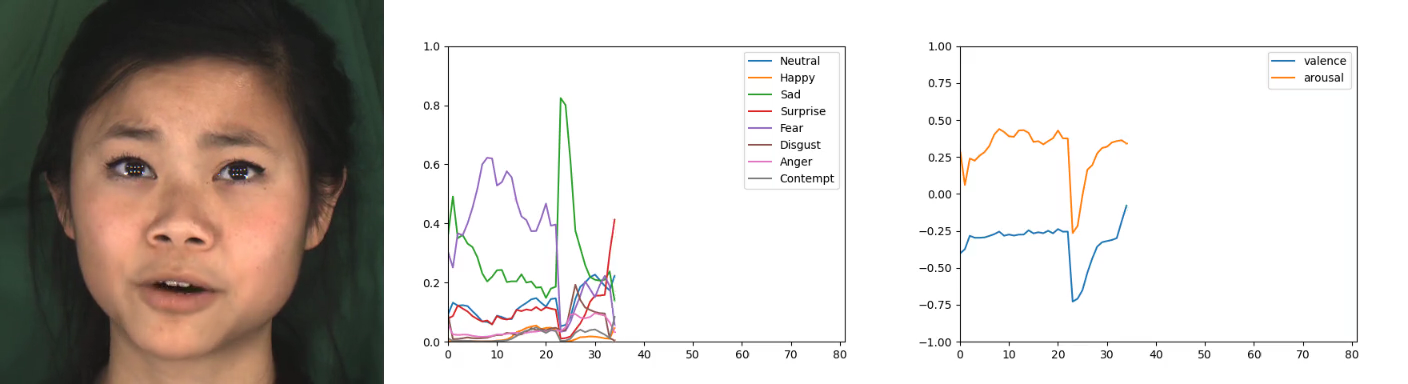}
    \includegraphics[width=1.\columnwidth]{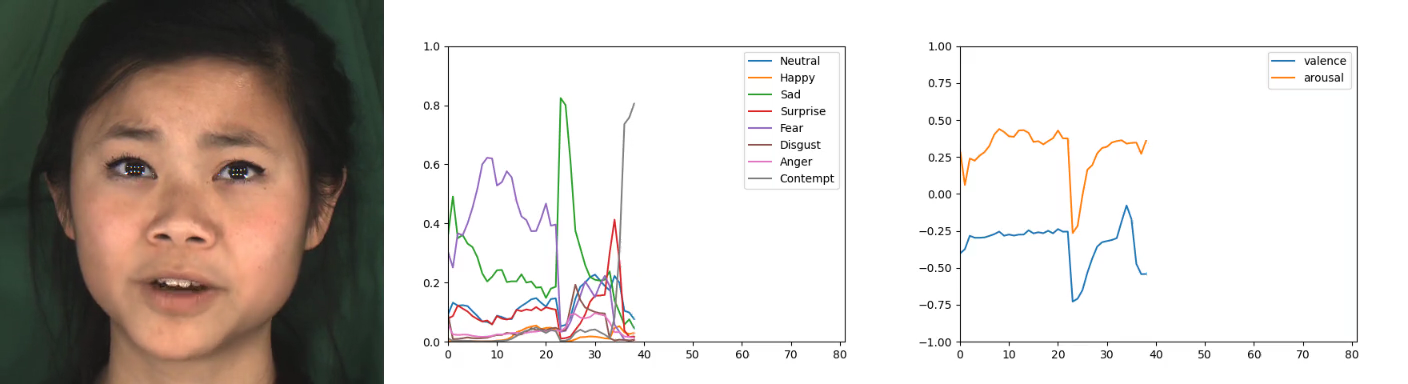}
    \includegraphics[width=1.\columnwidth]{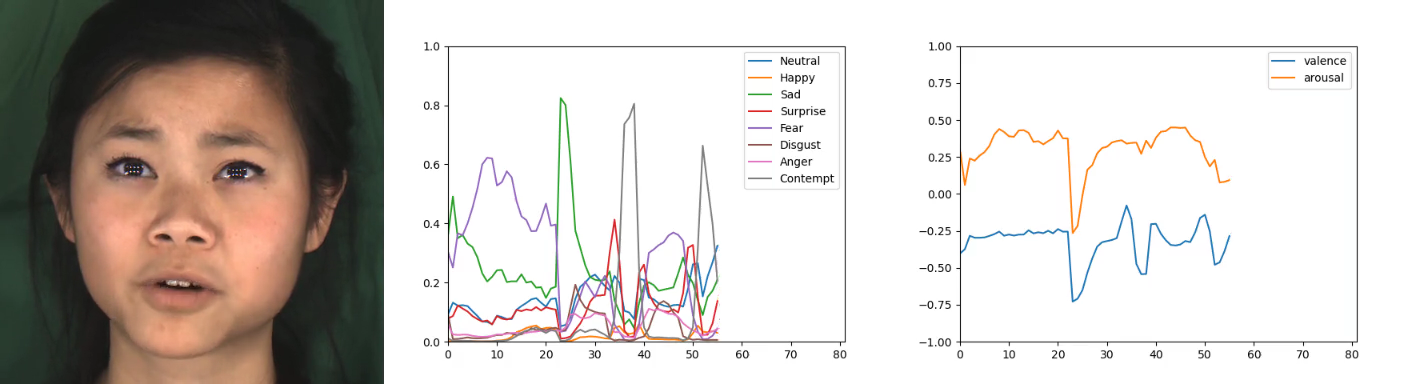}
    \includegraphics[width=1.\columnwidth]{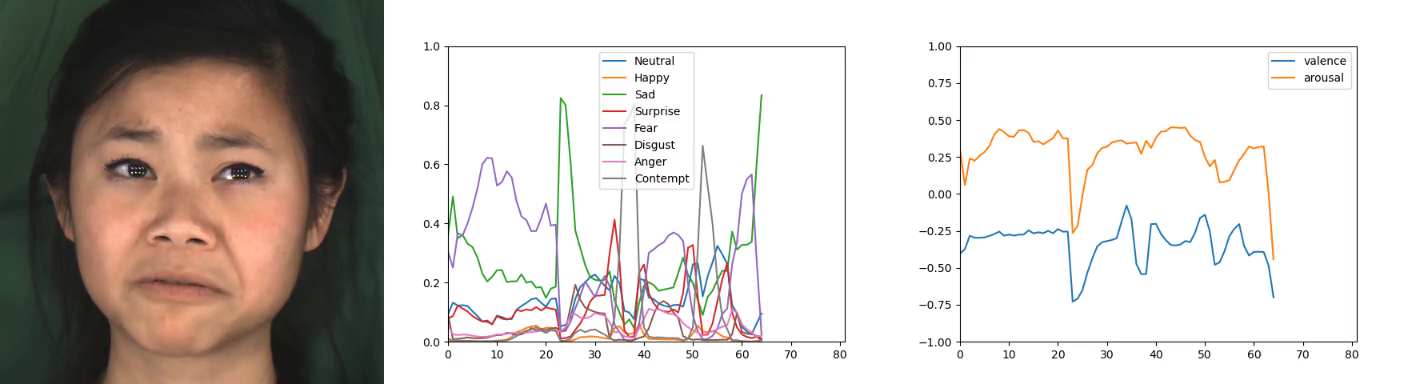}
    \caption{
        \textbf{Static Emotion Recognition}. In the figure you can see the input images taken out of one of the videos of a sad person from the MEAD dataset (left-hand side) and the outputs of a static emotion recognition system - the basic expression classification (middle) and valence and arousal (right-hand side) over the course of a video, with x-axis representing the temporal dimensions and y-axis the probability of the emotion class (middle) and valence, arousal (right).
        The person has a constant level of sadness throughout the video. 
        Despite that, the static emotion recognition yields very different classification for individual frames (fear, sadness, surprise, contempt and neutral).
    }
    \label{fig:static_emo_rec}
\end{figure*}

Static (single-image) emotion recognition is not stable. 
It can output many different classification results over the course of a video of a person talking under a single emotion as shown in Fig.~\ref{fig:static_emo_rec}.
Even to a human, single frames can be misleading because temporal context is missing and speech-induced facial expressions can lead to misinterpretation. 
\todelete{Automatic}\camrdy{Static} emotion recognition suffers from the same limitation. 
This limitation can be lifted, when considering emotions as a temporal phenomenon.
For this reason, we opt to train a video emotion recognition network that is able to leverage contextual information.
And then utilize the sequence aggregated features for our video emotion perceptual loss.

\subsection{Video Emotion Recognition}

\begin{figure*}[t]
    \offinterlineskip
    \centering
    \includegraphics[width=1.99\columnwidth]{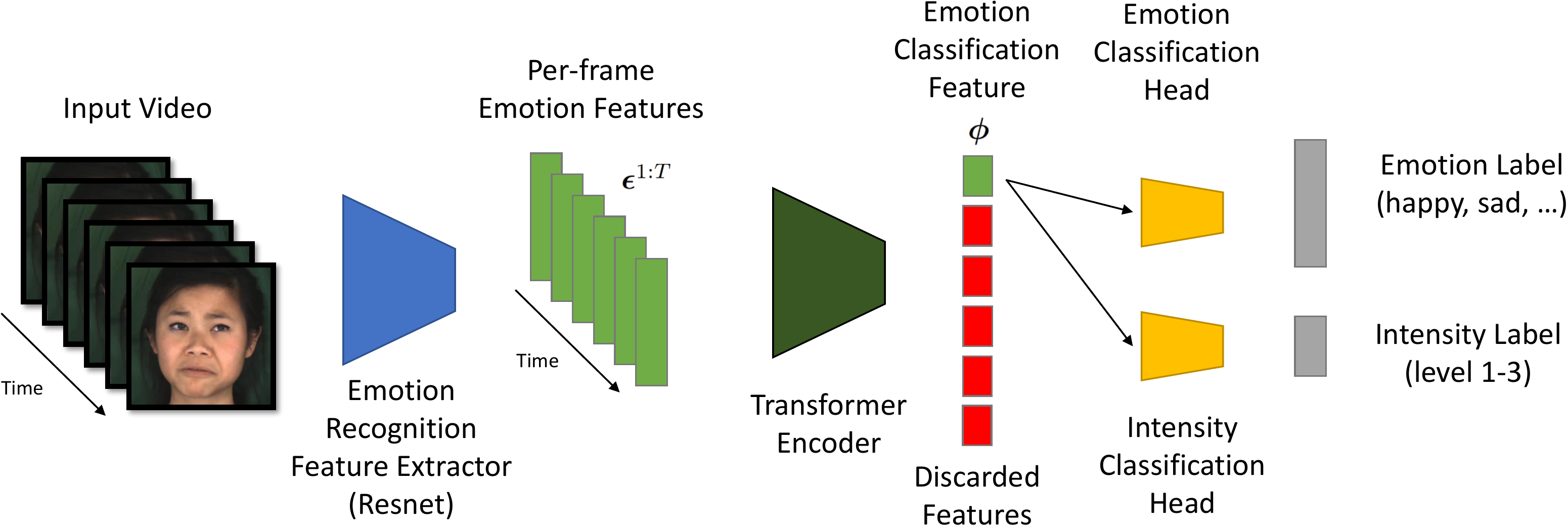}
    \caption{
        \textbf{Video Emotion Recognition Architecture.}
        The input video is passed frame by frame into the the pretrained and frozen emotion recognition net of Danecek et al.~\cite{EMOCA:CVPR:2021} obtaining a $2048d$ emotion feature for each frame $\emovec\tmp$.  
        These are passed into a single-layer transformer encoder to extract a single $256d$ video emotion feature $\videmovec$.  
        Instead of additive positional encodings we opt to use the ALiBi \cite{press2021train} mechanism and we modify it, such that it also considers future frames.
        This feature is then fed into linear classification heads to classify the emotion and the intensity. 
    }
    \label{fig:vid_emo_rec}
\end{figure*}
Our implementation of the video emotion networks is a lightweight single-layer transformer network with two classification prediction heads - one to classify the emotion and one to classify the intensity. 
It takes a sequence of static emotion recognition features extracted from a video on the input $\emovec\tmp$, passes it through a transformer encoder to get the video emotion feature $\videmovec$. The video features is then used to classify the emotion class and intesity with linear classification prediction heads.
We train this network on the MEAD dataset with the standard cross-entropy classification losses for both classification tasks (emotion and intensity) until convergence. 
Ground truth labels for both emotion and intensity are provided with the MEAD dataset.
The architecture is depicted in Fig.~\ref{fig:vid_emo_rec}.
The model code and weights will be made publicly available.

\subsection{Dynamic vs Static Emotion Loss}
Fig. \ref{fig:static_vs_dynamic} compares the performance of the static and dynamic emotion losses. 
The video emotion classification is significantly superior to the static emotion classifier.

\begin{figure*}[t]
    \offinterlineskip
    \centering
    \includegraphics[width=1.\columnwidth]{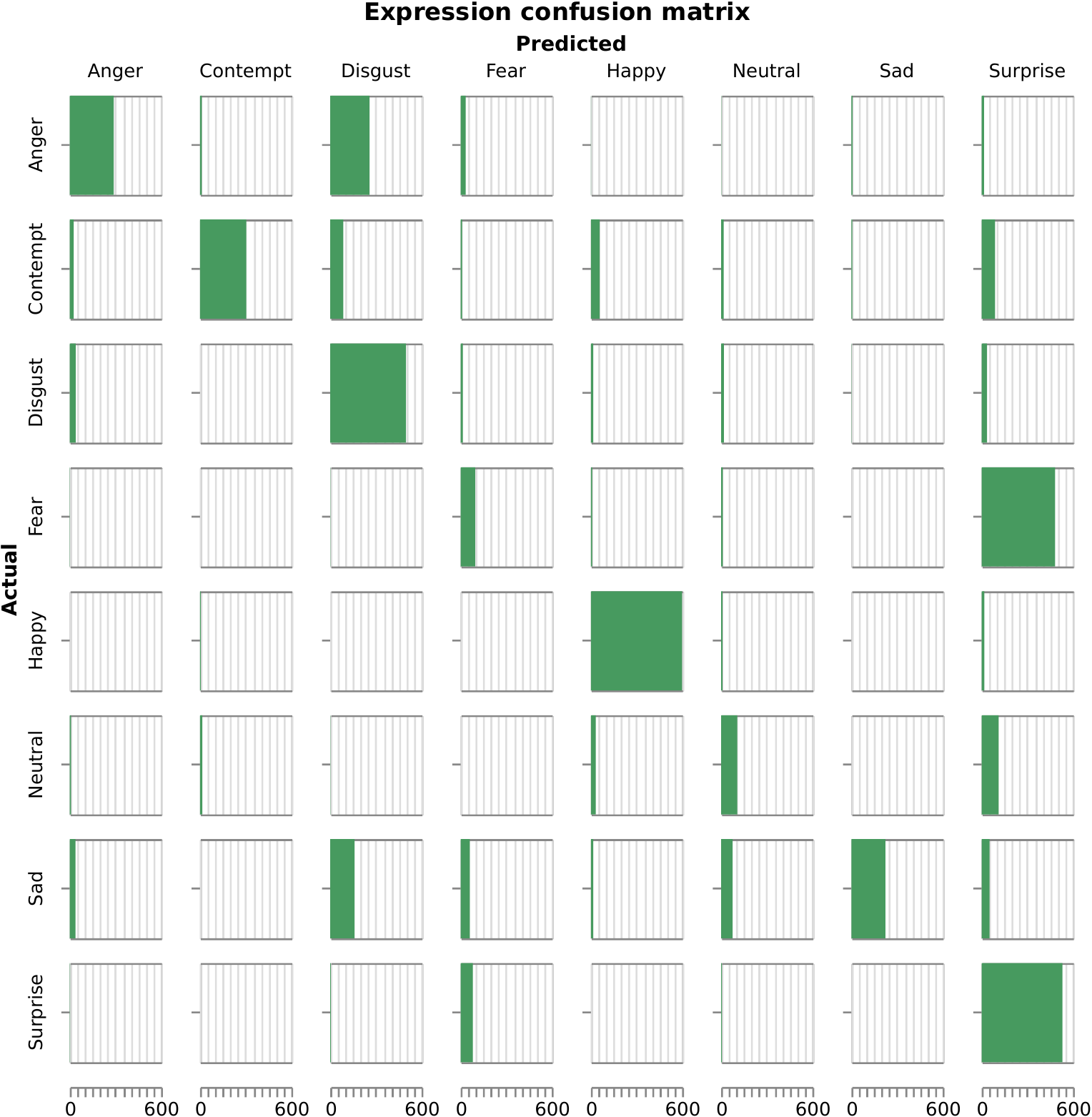}
    \includegraphics[width=1.\columnwidth]{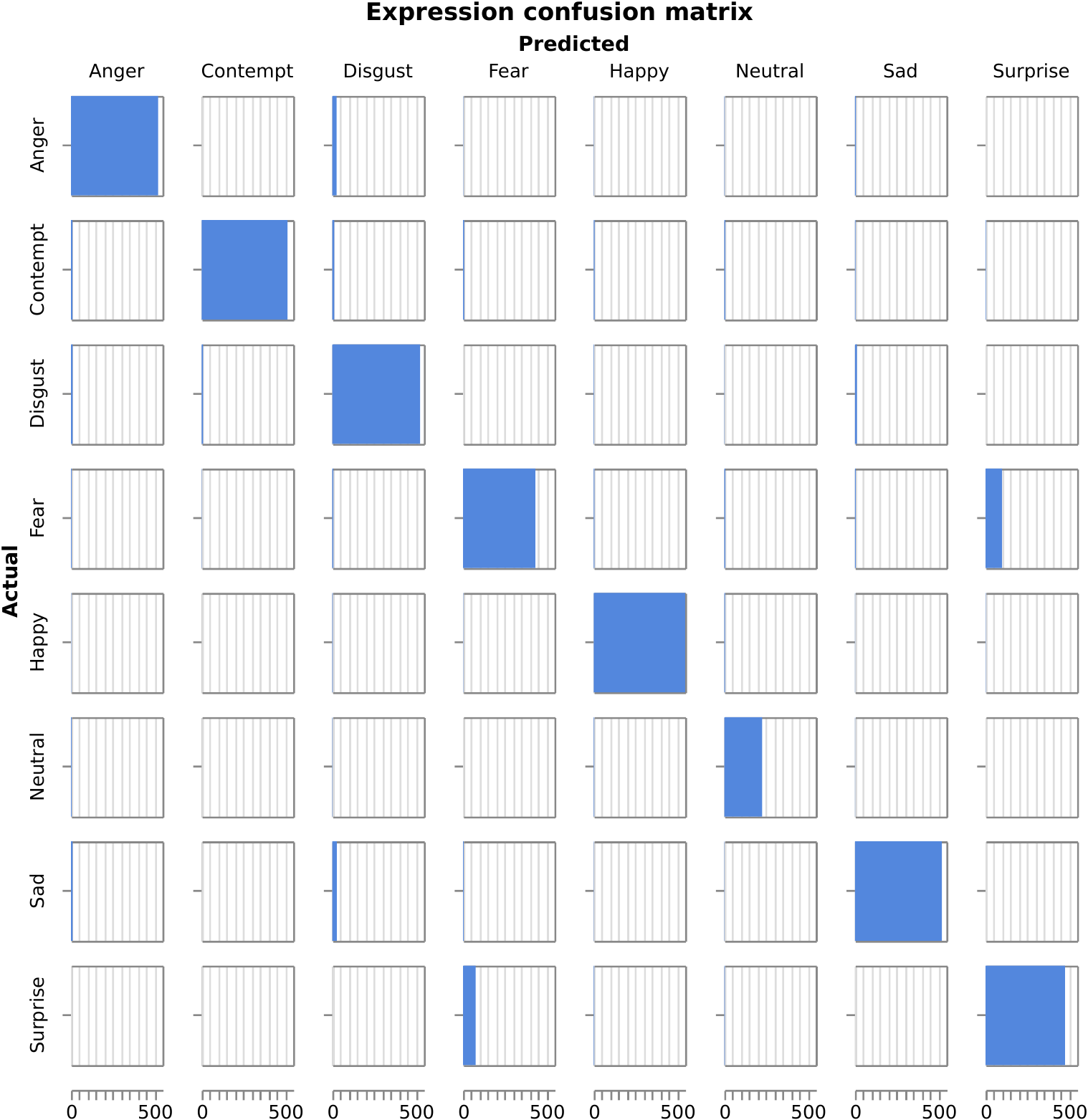}
    \caption{
        \textbf{Classification performance of static vs dynamic emotion loss}. Here we show two confusion matrices on the MEAD validation set, the one of static emotion recognition network (left) and a dynamic emotion recognition network (right). For static emotion classification results we took the most occurring classification in each video. The accuracy of static emotion recognition network is 57.9 \% and the dynamic is 90.8\%.
    }
    \label{fig:static_vs_dynamic}
\end{figure*}

\section{Datasets}

\begin{table}[b]
\centering
\caption{{\bf Datasets}}
\resizebox{0.49\textwidth}{!}{
\begin{tabular}{l|ccccc}
\toprule
    Datasets               &   Modality       &  Number of subjects &  Expressions &  Duration   \\
\midrule
    BIWI   &   3D                         &  14 (8F, 6M)             &  11                       &  1.43 h    \\
    VOCASET   &   3D                         &  12 (6F, 6M)              &  --                       &  0.48 h     \\
     S3DFM   &   3D                         &  77 (27F, 50M)              &  --                       &   0.28 h     \\
    Multiface      &   3D                         &  13             &  65 (v1), 118 (v2)                       &  --    \\
   VoxCeleb1         &   2D                         &  1251 (561F, 690M)             &  --                &  352 h     \\
    VoxCeleb2     &   2D                         &  6112 (2351F, 3761M)             &  --                 &  2442 h     \\
    Faceforensics++            &   2D                         &  --             &  --                &  --        \\
    CelebV-HQ         &   2D                         &  15653             &  83                    &  68 h      \\
    LRS2-BBC            &   2D                         &  --             &  --                    &  224.5 h     \\
    LRS3-TED            &   2D                         &  5594             &  --                      &  438 h     \\
    RAVDESS            &   2D                         &  24 (12F, 12M)             &  8                    &  --    \\
    CREMA-D            &   2D                         &  91 (43F, 48M)            &  6                     &  --      \\
    MELD            &   2D                         &  407             &  7 (+ 3 sentiments)                     &  12.96 h     \\
    CMU-MOSI            &   2D                         &  98             &  sentiment intensity [-3,3]                   &  2.6 h       \\
    CMU-MOSEI            &   2D                         &  1000             &  6 (+ 5 sentiments)                    &  65.88 h        \\
    TalkingHead-1KH            &   2D                         &  --             &  --                &  1000 h         \\
    MEAD            &   2D                         &  60 (30F, 30M)             &  8                     &  38.95 h    \\
    
\bottomrule
\end{tabular}
}
\label{table:data}
\end{table}

\paragraph{Existing datasets:}
Table~\ref{table:data} provides an overview of existing 3D and 2D face datasets with synchronized audio. 
While there is seemingly many datasets available, each of them come with a particular set of challenges (such \camrdy{as} low video quality, too difficult for face reconstruction, not enough variety of emotions and speaking styles, small size of the dataset etc.). 
Taking all of the above into account, we opt for using MEAD~\cite{kaisiyuan2020mead} in our experiments, since it is of sufficient scale, has all emotions in different intensities \camrdy{and thanks to its constrained environment and high quality video, it is relatively easy to reconstruct.}
It is also considerably easier to reconstruct compared to the other datasets since it is a captured in a lab and hence does not come with the problems of associated with reconstruction in-the-wild (such as occlusions, difficult illumination etc.).

\section{Perceptual studies}

Here we provide additional detail about our perceptual studies. 

\paragraph{Lip articulation evaluation:}
Fig.~\ref{fig:LipSyncCompare} show the exact layout used for this perceptual study. 

\begin{figure*}[t]
    \offinterlineskip
    \centering
    \includegraphics[width=1.9\columnwidth]{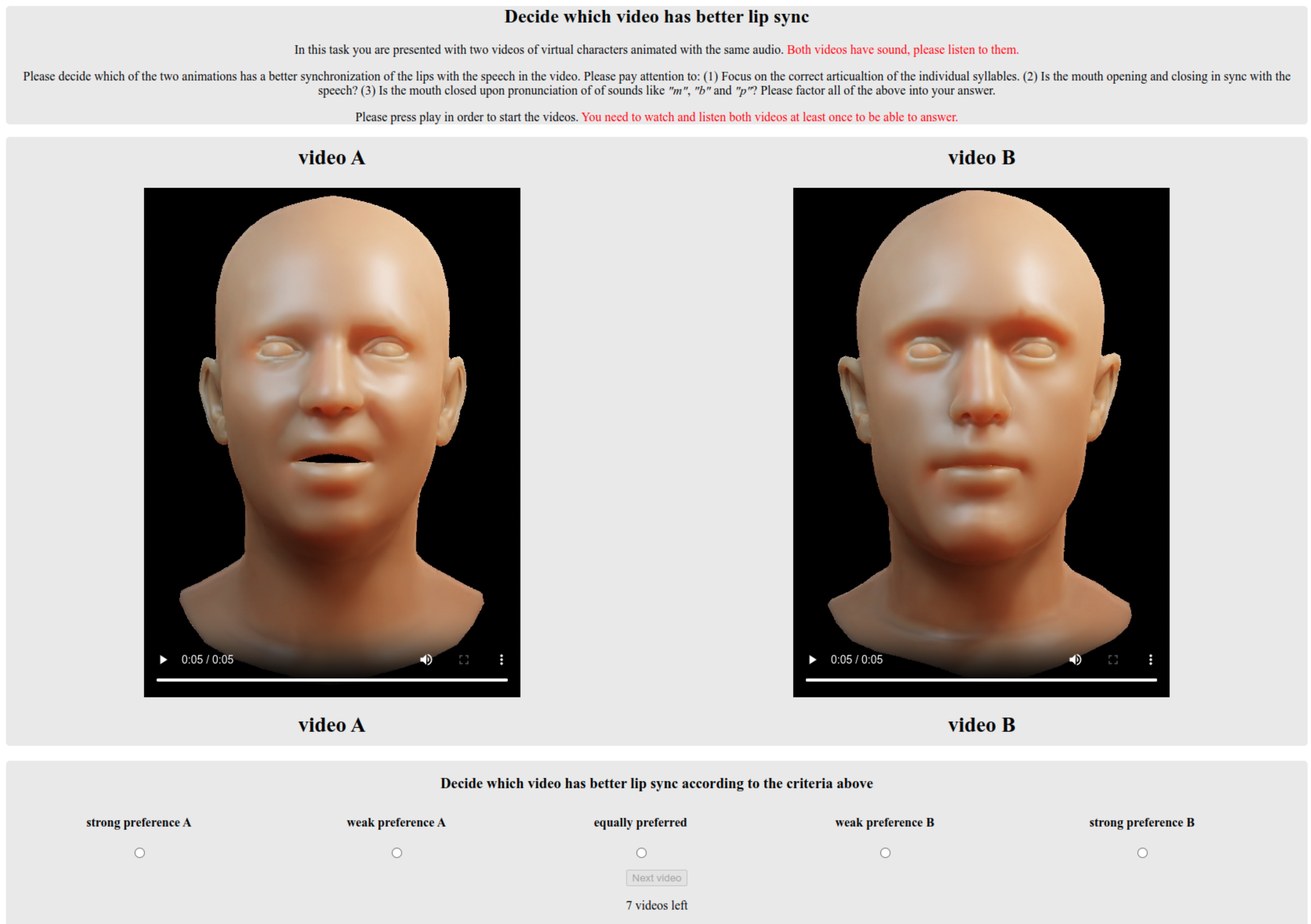}
    \caption{
        \textbf{Lip-sync Comparison with SOTA}. The participants are presented with two videos \textit{with audio}, one generated using \model and others using SOTA methods. The participant is asked to judge the quality of synchronization of the lips with the speech in the video. 
    }
    \label{fig:LipSyncCompare}
\end{figure*}

\paragraph{Ablation perceptual study:}
Figures \ref{fig:EmotionAnalyze} and \ref{fig:LipSyncAnalyze} show the web layout of the emotion quality and lip-sync quality ablation studies respectively.  
The two layouts alternate after each response. 

\begin{figure*}[t]
    \offinterlineskip
    \centering
    \includegraphics[width=1.9\columnwidth]{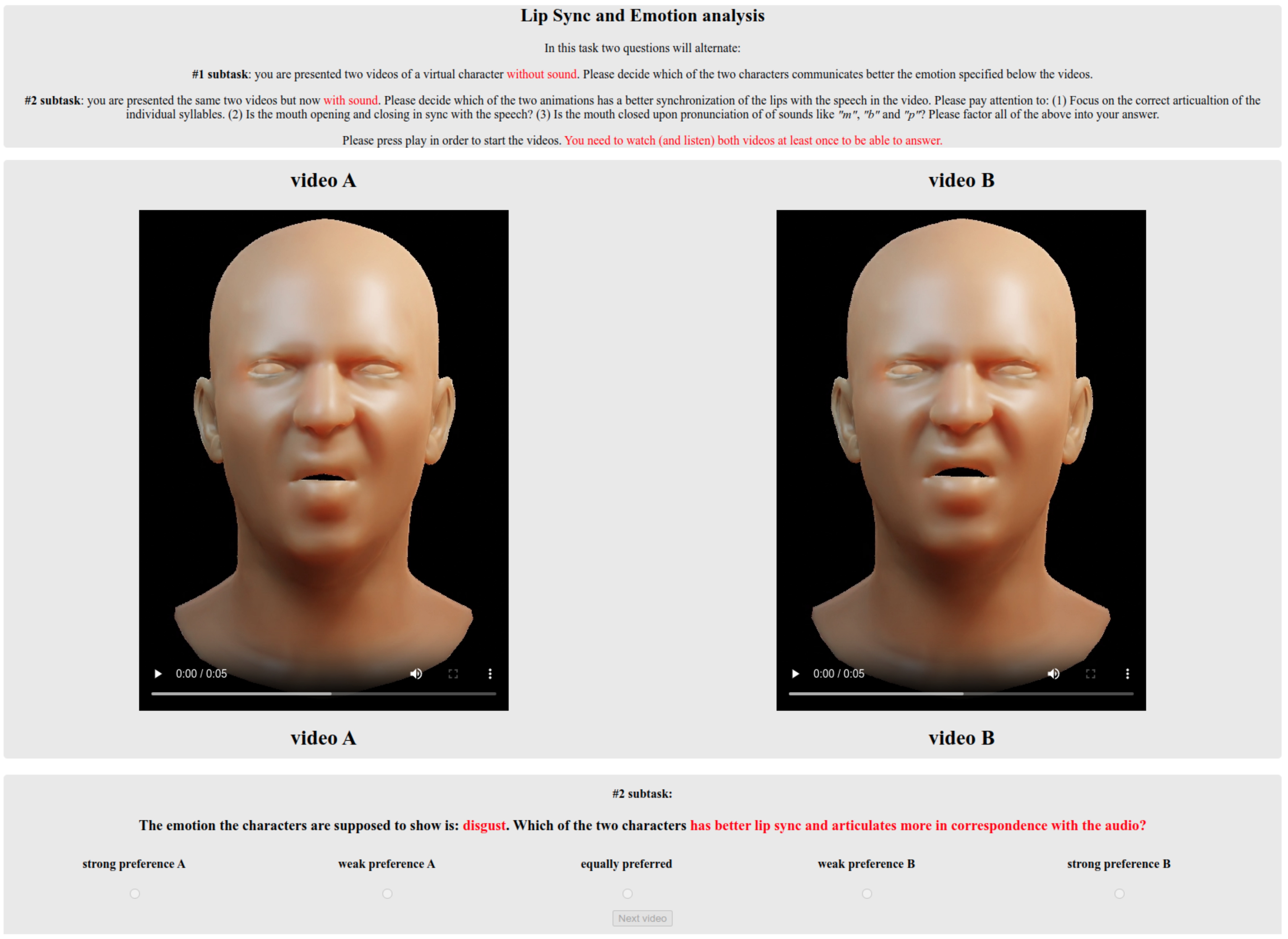}
    \caption{
        \textbf{User study: Lip-sync Analysis}. The participants are presented with same two videos as in Fig.~\ref{fig:EmotionAnalyze}. This time the videos are \textit{audible} and the user is asked to judge the quality of the articulation, taking the audio into account. Again, the participant must watch both videos in full length before being able to proceed to the next question. Upon answering, the user is redirected to the first question (emotion quality assessment) with a new pair of videos.
    }
    \label{fig:LipSyncAnalyze}
\end{figure*}

\begin{figure*}[t]
    \offinterlineskip
    \centering
    \includegraphics[width=1.9\columnwidth]{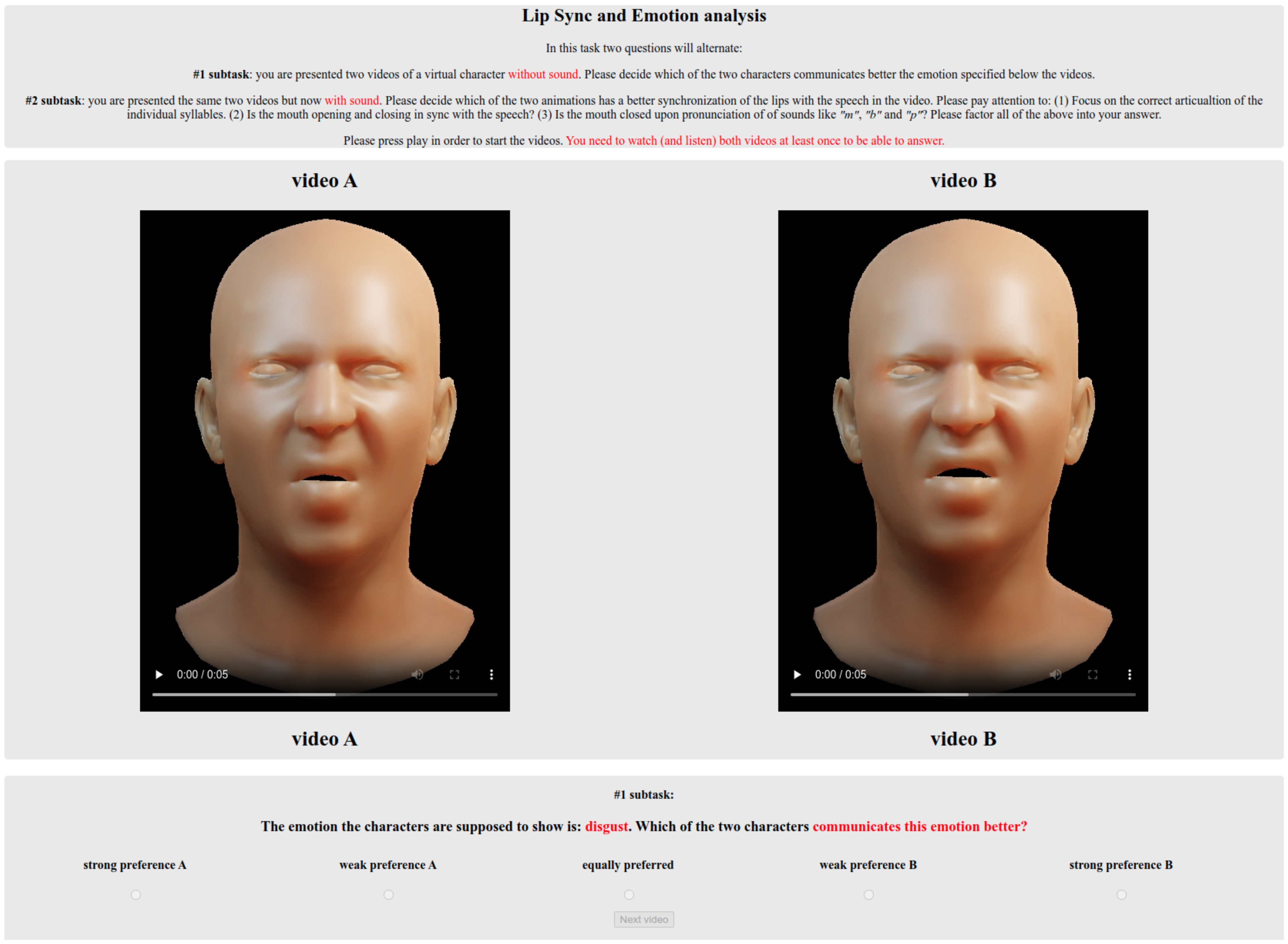}
    \caption{
        \textbf{Emotion quality assessment.} The participants are presented with two \textit{muted} videos and are asked to select which of the two videos better communicates the emotion specified in the text under the video. The participant must watch both videos in full length at least once before submitting an answer becomes available and then they proceed to the next question about lip-sync assessment for the same two videos (see Fig.~\ref{fig:LipSyncAnalyze}).
    }
    \label{fig:EmotionAnalyze}
\end{figure*}

\section{Additional Results}
Fig.~\ref{fig:additional_results} and Fig.~\ref{fig:additional_results2} demonstrates the ability of \model to generate emotional animation. A different speaking style was used for each of the two figures.

\begin{figure*}[t]
    \offinterlineskip
    \centering
    \includegraphics[width=1.99\columnwidth]{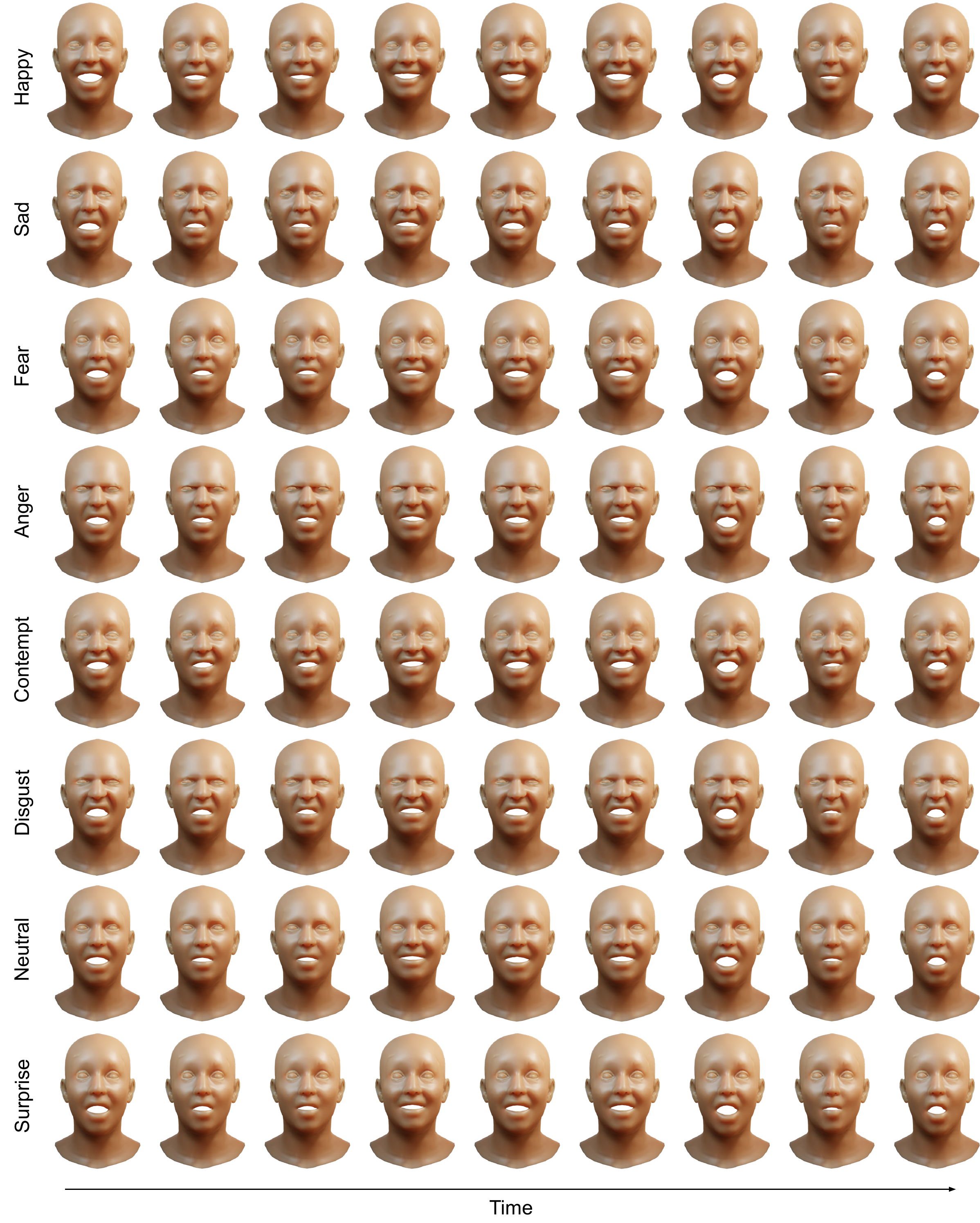}
    \caption{
        \textbf{Additional results with the same input audio but different input emotion.}
    }
    \label{fig:additional_results}
\end{figure*}

\begin{figure*}[t]
    \offinterlineskip
    \centering
    \includegraphics[width=1.99\columnwidth]{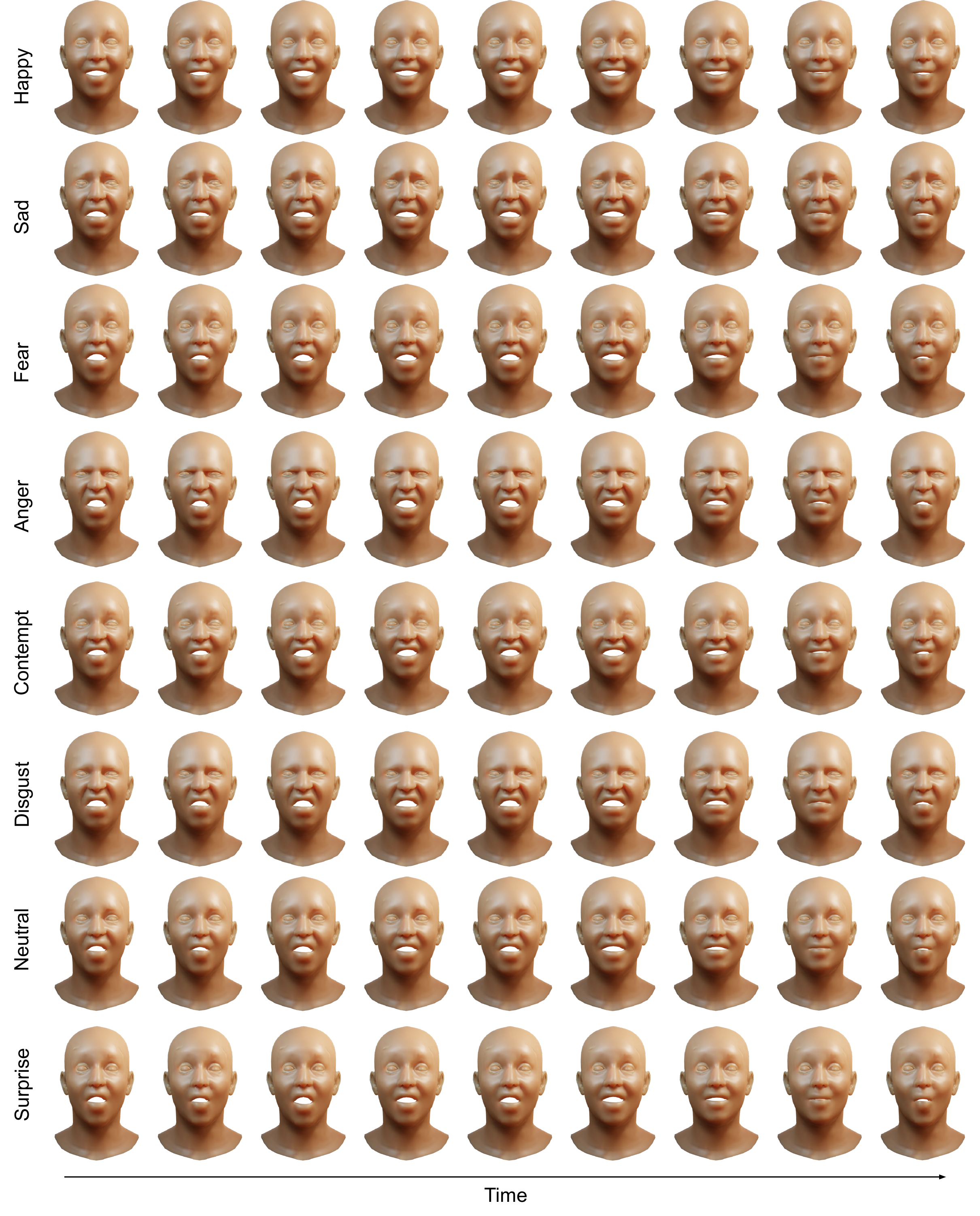}
    \caption{
        \textbf{Additional results with the same input audio but different input emotion.}
    }
    \label{fig:additional_results2}
\end{figure*}



 \fi

\end{document}